%% file: main.tex
\documentclass{article}

\usepackage[nonatbib,preprint]{neurips_2023}

\usepackage[numbers]{natbib}

\usepackage[utf8]{inputenc} % allow utf-8 input
\usepackage[T1]{fontenc}    % use 8-bit T1 fonts
\usepackage{hyperref}       % hyperlinks
\usepackage{url}            % simple URL typesetting
\usepackage{booktabs}       % professional-quality tables
\usepackage{amsfonts}       % blackboard math symbols
\usepackage{nicefrac}       % compact symbols for 1/2, etc.
\usepackage{microtype}      % microtypography
\usepackage{multirow, makecell}

\usepackage{wrapfig}

\usepackage{float}
\usepackage{amsmath}

\usepackage{graphicx}
\usepackage{caption}
\usepackage{subcaption}
\usepackage{xcolor}

\definecolor{dartmouthgreen}{rgb}{0.05, 0.5, 0.06}
\definecolor{deepcarmine}{rgb}{0.66, 0.13, 0.24}
\definecolor{d4color}{rgb}{0.9686274509803922, 0.7137254901960784, 0.8235294117647058}

\title{D4: Improving LLM Pretraining via Document De-Duplication and Diversification}

\author{%
  Kushal Tirumala* \\
  Meta AI Research\\
  \And
  Daniel Simig*\\
  Meta AI Research \\
  \And
  Armen Aghajanyan \\
  Meta AI Research \\
  \And
  Ari S. Morcos \\
  Meta AI Research \\
}

\begin{document}

\newcommand{\todo}[1]{\textcolor{red}{[TODO: #1]}}
\def\thefootnote{*}\footnotetext{Equal contribution. Correspondence emails: ktirumala@meta.com, simigd@gmail.com}\def\thefootnote{\arabic{footnote}}
\maketitle

\input{0_abstract}
\input{1_introduction}

\input{5_related_work}
\input{2_experimental_setup}
\input{3_results}

\input{6_limitations}

\section{Acknowledgements}
The authors would like to thank many people who helped bring this work to fruition: Srini Iyer, Yuchen Zhang, Todor Mihaylov, Jacob Xu  Moya Chen, Mansheej Paul, Mitchell Wortsman, Amro Abbas, Aaditya Singh, Myra Cheng, and Matthew Leavitt. The authors would also like to thank Surya Ganguli, Mona Diab, and Xian Li for initial brainstorming and are grateful for help with compute infrastructure given by Henry Estela and Victoria Lin. Lastly, the authors would like to thank anonymous reviewers for improving the quality and writing of this paper.

\bibliographystyle{plainnat}
\bibliography{citation}

\newpage
\appendix
\input{appendix}

\end{document}

%% file: 0_abstract.tex
\begin{abstract}
% v2
Over recent years, an increasing amount of compute and data has been poured into training large language models (LLMs), usually by doing one-pass learning on as many tokens as possible randomly selected from large-scale web corpora. While training on ever-larger portions of the internet leads to consistent performance improvements, the size of these improvements diminishes with scale, and there has been little work exploring the effect of data selection on pre-training and downstream performance beyond simple de-duplication methods such as MinHash. Here, we show that careful data selection (on top of de-duplicated data) via pre-trained model embeddings can speed up training (20\% efficiency gains) and improves average downstream accuracy on 16 NLP tasks (up to 2\%) at the 6.7B model scale. Furthermore, we show that repeating data intelligently consistently \textit{outperforms} baseline training (while repeating random data performs worse than baseline training). Our results indicate that clever data selection can significantly improve LLM pre-training, calls into question the common practice of training for a single epoch on as much data as possible, and demonstrates a path to keep improving our models past the limits of randomly sampling web data. 
\end{abstract}

%% file: 1_introduction.tex
\section{Introduction}

Due to computational limits, initial work on language model pre-training focused on training models on small, high-quality text datasets such as BookCorpus \citep{zhu2015aligning} and Wikipedia \citep{merity2016pointer}. More recently, however, catalyzed by works like \citep{Radford2019LanguageMA}, advancements in large language models (LLMs) have been driven by leveraging large collections of unlabeled, uncurated data derived from snapshots of the internet (CommonCrawl \citep{ Gao2020ThePA, penedorefinedweb, raffel2020exploring}), trading off small quantities of heavily-curated data for huge quantities of less-curated data. Because of the dramatic increase in data quantity, these strategies have resulted in higher performance models and have sparked a new paradigm wherein massive, largely unfiltered datasets are utilized for training \citep{chowdhery2022palm, smith2022using, Touvron2023LLaMAOA}.

Despite the essential role that large-scale web data now play in LM pre-training, data curation and selection for large-scale web data have not been thoroughly explored. This is primarily due to the universality of compute and data scaling laws \cite{Hoffmann2022TrainingCL, Kaplan2020ScalingLF} which give practitioners a low-risk way to reliably improve LM performance by merely adding ``more'' data, not necessarily the ``right'' data. Indeed, the data selection method used to model scaling laws (along with the data selection methods used in most LLM pre-training pipelines) involves simply randomly sampling tokens from web data dumps that have been put through a combination of simple heuristic filtering (e.g., to eliminate very short strings) and very near match de-duplication \citep{Lee2021DeduplicatingTD}. 

If we continue relying on scaling laws to improve LLMs, we will quickly hit diminishing returns due to the power-law nature of scaling laws. We will therefore need exponentially more data to maintain a consistent marginal improvement, which may prove especially challenging as we are fast approaching the limits of available human-generated text data \citep{villalobos2022run}. Encouragingly, in the context of vision, \citet{Sorscher2022BeyondNS} demonstrated that we could leverage simple data selection strategies to overcome costly power-law scaling. They compare numerous data selection methods and find that clustering data points in a pre-trained embedding space and ranking according to the distance to the cluster centroid ("SSL Prototypes") significantly  improves the data efficiency of vision models. Recently, \citet{Abbas2023SemDeDupDL} demonstrated that using a pre-trained embedding space to de-duplicate data ("SemDeDup") improves both efficiency and performance of vision-language models such as CLIP. However, there has been little exploration of these or related approaches in training LLMs at scale. Motivated by this, we argue that by combining these approaches and applying them to LLMs, relatively simple data selection strategies leveraging pre-trained embeddings can significantly improve LLM training. Specifically, our contributions are as follows:

\begin{itemize}
    \item We investigate different data selection strategies for standard LLM pre-training setups where data has already been manually filtered / de-duplicated (e.g., MinHash), and where we do not know the target distribution for which we optimize performance. We argue that the performance of SSL Prototypes is affected by duplicate-driven clusters in the embedding space. In Section~\ref{sec:methods} we propose a new data selection strategy \textbf{D4} that utilizes SemDeDup to avoid getting impacted by such clusters.

    \item In Section~\ref{sec:results_fixed_compute}, we show that in the \textit{compute-limited regime} where we have “infinite” source data and train models with fixed token budgets, we can achieve better pre-training perplexity and downstream accuracy than random iid data selection and previously established methods. Furthermore, we show that our method D4 can achieve around 20\% efficiency gains at the 6.7b model scale, and that the magnitude of efficiency gains increases with model scale.
    \item In the \textit{data-limited regime}, where we run out of data and must epoch over data, cleverly choosing what data to repeat can beat training on randomly selected new data, whereas randomly choosing data to repeat underperforms adding new data (Section~\ref{sec:results_repeated_data}). This calls into question the standard practice of single epoch LLM training, and suggests that epoching over intelligently subselected data might be a better approach.
\end{itemize}

\begin{figure}[h]
\begin{center}
\includegraphics[width = 1.0\textwidth]{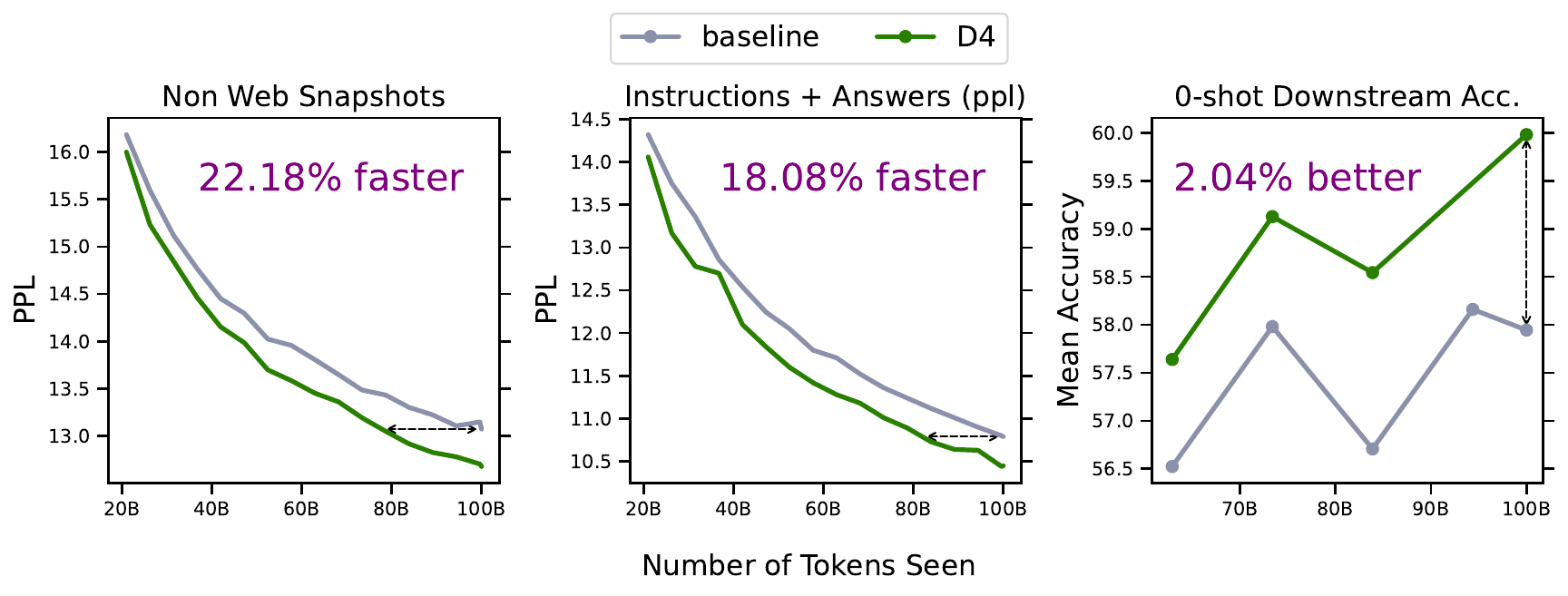}

\caption{Learning curves for 6.7B OPT model pretraining on 100B tokens, with data selected with D4 (pink line) and randomly (gray line). D4 significantly outperforms baseline training, getting between 18-20\% efficiency gains on validation perplexity and 2\% increase in average 0-shot downstream accuracy across 16 NLP tasks.  See Section~\ref{sec:efficiency_gains_across_model_scales_and_training} for full learning curves.}
\vspace{-1em}
\label{fig:6.7b_panel_learning_curve}
\end{center}
\end{figure}

%% file: 5_related_work.tex
\section{Related Work}
\textbf{Data selection in non-text domains:} Numerous works have successfully used data selection techniques in vision models \citep{birodkar2019semantic, chitta2021training, jiang2019accelerating, meding2021trivial, mindermann2022prioritized, paul2021deep, toneva2018empirical}, though these have largely been at sub-ImageNet scale. Some of these works develop pruning metrics that score individual data points (for example, EL2N from \citet{paul2021deep}), while some focus on data-efficiency and attempt to find groups of points that allow models to reach baseline performance with less data points, e.g., coresets \citep{cazenavette2022dataset, mirzasoleiman2020coresets, sener2017active, zhao2020dataset}. \citet{Sorscher2022BeyondNS} compares many of the existing individual-score methods at ImageNet scale, finding that their SSL prototypes metrics and the (prohibitively expensive) memorization metric from \citet{feldman2020neural} generally outperforms other methods. In the audio domain, \citet{dong2020audio} computes importance embeddings to find important training samples for audio scene classification. More recently, \citet{Abbas2023SemDeDupDL} demonstrated very encouraging results on vision-language models (CLIP models) using SemDeDup — a similar method to SSL prototypes but focused on semantic deduplication.  Our work combines these approaches and applies them to large-scale LLMs.

\textbf{Effect of pre-training data on LM performance:} \citet{Gao2020ThePA} trains variants of GPT-2 \citep{Radford2019LanguageMA} models from scratch to compare the "Pile" dataset to CommonCrawl-derived corpora. \citet{Radford2019LanguageMA} demonstrates the positive impact of the quality filters and data de-duplication methods used to curate MassiveWeb by training 1.4B parameter models from scratch. \citet{Hernandez2022ScalingLA} quantifies the effect of various amounts of artificially created data duplication and provides analysis on interpreting the changes in the behaviour of the models trained on duplicated data. Concurrently to our work, \citet{Xie2023DataSF} propose using importance resampling to align the distribution of web data to high-quality reference corpora such as Wikipedia. Similarly, \citet{gururangan2020don} explores data selection strategies for adapting LMs to a task-specific corpus. Another line of recent work explores how data mixture affects pre-training, with \citet{Xie2023DoReMiOD} demonstrating impressive improvements in downstream accuracy and perplexity across all datasets for 8B parameter models trained on the Pile. Similarly, \citet{Longpre2023APG} explores the role of text quality, toxicity, age, and domain distribution of training data on LLM performance. Outside of data curation, there has been a recent surge of work exploring the impact of repeating data \citep{biderman2023pythia, Muennighoff2023ScalingDL, xue2023repeat}, generally concluding that repeating tokens is worse than training on new tokens (which we question in Section~\ref{sec:results_repeated_data}).

%% file: 2_experimental_setup.tex
\section{Experimental Setup}

\paragraph{Notation} Given a source dataset, $D_{source}$, of documents (crawled web pages) and model architecture, $M$, we aim to find a strategy $S$ for selecting a subset of these documents  that maximizes some evaluation metric $E(M(D_{S, R}))$. $R$ indicates the proportion of remaining documents from the source dataset $D_{source}$ after selecting data with strategy $S$. For this reason, we refer to $R$ throughout this work as the \textit{selection ratio}: for example, if $R = 0.25$ and $|D_{source}| = 100$ million, then we \textit{select} 25\% of documents from a source dataset of size $100$M documents to arrive at a a training dataset with $25$M documents. We operate at the granularity of a single document, independently of how the model trainer would pack these documents into batches later. Throughout the paper, we use random selection as the baseline for $S$, as it is the most common method for selecting data for language model pre-training. In the rest of this section, we describe our choices of source dataset ($D_{source}$), model ($M$), evaluation metric ($E$), and, most importantly, our suggestions for the selection strategy ($S$).

\subsection{Training Dataset (choice for $D_{source}$)}
We perform all of our training runs on a version of CommonCrawl pre-processed with a CCNet \citep{Wenzek2019CCNetEH} pipeline identical to the one used by \citet{Touvron2023LLaMAOA}. We add an additional step of MinHash-based de-duplication (see more details in Section~\ref{sec:experimental_setup_details}). Applying this common step before our experiments guarantees that any effects observed in our experiments complement the currently prevalent approach of MinHash-based data de-duplication strategies. Throughout the rest of this work, we refer to this dataset as \textit{CC-dedup}. 

\subsection{Model Training (choices for $M$ and $T_{target}$)}
To evaluate different configurations of data selection strategies, we train OPT \citep{Zhang2022OPTOP} models from scratch on the pruned versions of datasets. We use the standard model architectures and settings of \citet{Zhang2022OPTOP} and use MetaSeq \citep{Zhang2022OPTOP} to train all our models. For 125M models, we train to $T_{target} = 3B$ tokens. For 1.3B parameter models, we train to target token count of $T_{target} = 40B$. For 6.7B parameter models, we train to $T_{target} = 100B$ tokens. We choose these by trimming down the token budgets suggested by \citet{Hoffmann2022TrainingCL} to meet our compute limitations. We provide full details of our training setup in Section~\ref{sec:experimental_setup_details}.

\subsection{Evaluation Metrics (choices for $E$)}

We keep most of our evaluation consistent with the setup from \citet{Zhang2022OPTOP}.

\textbf{Validation Set Perplexity}. Our validation sets mainly come from \cite{Zhang2022OPTOP}, which includes validation sets derived from subsets of the Pile \cite{Gao2020ThePA} such as CommonCrawl, DM Mathematics, HackerNews, OpenSubtitles, OpenWebText2, Project Gutenberg, USPTO, Wikipedia. We also include a validation set obtained from the PushShift.io Reddit dataset \citep{baumgartner2020pushshift} (which we refer to as \textit{redditflattened}). In addition, we measure perplexity on a validation set obtained from a train-validation split of our source dataset \textit{CC-dedup}, and a validation set from C4 \cite{raffel2020exploring}.

We notice that the effects of data selection vary significantly on individual validation sets depending on whether the validation set was derived from a web data corpus or not (see more details and analysis in Section \ref{sec:web_analysis}). Motivated by this, we split validation sets into Web-snapshots (C4, CommonCrawl, and CC-dedup) and Non-web snapshots, and report average perplexity within these sets.

\textbf{Downstream Task Accuracy.} To evaluate downstream performance of our trained models, we report average 0-shot accuracy across the 16 NLP tasks from \citet{Zhang2022OPTOP}, and use a prompting methodology consistent with \citet{Zhang2022OPTOP}. These set of 16 NLP tasks include Arc Challenge and ArcEasy \cite{clark2018think}, HellaSwag \cite{zellers2019hellaswag}, OpenBookQA \cite{mihaylov2018can}, PIQA \cite{bisk2020piqa}, StoryCloze \cite{mostafazadeh2016corpus}, Winograd \cite{levesque2012winograd}, Winogrande \cite{sakaguchi2021winogrande}, as well as tasks from SuperGLUE \cite{wang2019superglue}. We refer the reader to \citet{Zhang2022OPTOP} for more information about this evaluation setup.

\textbf{Instruction Tuning Perplexity}. The evaluation mentioned above metrics presents an inherent trade-off. Though accuracy on downstream tasks is typically viewed as a more concrete representation of a language model's real-world value, its variance tends to be higher due to the limited number of examples in these tasks and the step-wise behavior of accuracy as a metric. In contrast, perplexity, as a metric, is smoother while still exhibiting a strong correlation with performance \cite{schaeffer2023emergent}.
Therefore as a middle ground between the two evaluation metrics, we propose evaluating the perplexity on a sample drawn from the instruction-tuning dataset used for fine-tuning OPT-IML \cite{Iyer2022OPTIMLSL}. This dataset spans over 1500 unique NLP tasks and comprises a wide array of prompt-answer pairs and therefore is representative of the \textit{average} NLP task. It has been carefully crafted by merging extensive task collections such as Super-NaturalInstructions \cite{Wang2022SuperNaturalInstructionsGV} and PromptSource \cite{Bach2022PromptSourceAI}. We refer the reader to Table 2.1 in \cite{Iyer2022OPTIMLSL} for a comprehensive breakdown. This approach allows us to balance practical performance measures and statistical consistency in evaluation. We note that this metric can simply be considered as perplexity on another validation set, where the validation set is filled with examples used for instruction-tuning (we are \textbf{not} fine-tuning on this dataset).

\subsection{Data Selection Strategies (choices for $S$)}
\label{sec:methods}

In our initial exploration of un-curated web data, we embedded a large sample of web documents, clustered these embeddings, and manually inspected the resulting clusters. We quickly identified several high density clusters with documents that had little to do with the natural distribution of human language and were artifacts of the web crawling: for example, advertisements of Nike shoes that were automatically generated from a single underlying template with minor modifications (see Section~\ref{sec:appendix_examples_duplicate_driven_clusters} for details). 

Motivated by the intuition that these duplicate-driven clusters need tshould be pruned, as well as the recent success of pruning methods in vision and vision-language models \cite{Abbas2023SemDeDupDL, Sorscher2022BeyondNS}, we focus our efforts on data selection strategies that manipulate data points based on their position in an embedding space. We embed each document by feeding it into a 125M OPT model and use the last-layer embedding of the last token (we experiment with different embedding spaces in Section~\ref{sec:appendix_choice_of_embedding_space}). Following this, we experiment with several approaches:

\textbf{SemDeDup}: \citet{Abbas2023SemDeDupDL} proposed de-duplicating in both text and image domains by first using K-Means to cluster the embedding space, and removing points in each cluster that are within epsilon-balls of one another. We use this algorithm without any modifications and refer the reader to \citet{Abbas2023SemDeDupDL} for implementation details of this algorithm. 

\textbf{Prototypicality}: \citet{Sorscher2022BeyondNS} investigated a large variety of data pruning strategies to improve the data efficiency of training image classification models, including a newly introduced "SSL Prototypes" metric that proved to be one of their best methods. This strategy involves first clustering the embedding space using k-means clustering and discarding data points in increasing order of their distance to the nearest cluster centroid, such that the most "prototypical" data points are discarded, enriching the much higher variance outliers. We refer the reader to \citet{Sorscher2022BeyondNS} for a more detailed description of this algorithm.

\textbf{D4}: As mentioned previously, we find many instances of duplicate-driven clusters: clusters of templated text or extremely semantically redundant information that are not removed by MinHash. These regions of embedding space tend to be very dense and cause k-means to waste valuable cluster assignments on duplicated text. This biased clustering could also negatively to impact the effectiveness of SSL Prototypes since many clusters will be entirely driven by duplicates instead of more topical coherence. This insight lead us to our proposed strategy:

\begin{enumerate}
    \item Apply \textit{SemDeDup} with a selection ratio $R_{dedup}$ on the entire dataset $D$, producing a smaller dataset $D'$
    \item Cluster points in $D'$ with K-Means
    \item Apply \textit{SSL Prototypes} on $D'$, with a selection ratio $R_{proto}$
\end{enumerate}

The above-described strategy has an overall selection ratio of $R = R_{dedup} 
* R_{proto}$ and intends to diversify the distribution of our data locally and globally. For brevity we refer to this method as \textbf{D4}, a shorthand for \textit{Document De-Duplication and Diversification}. Throughout this work, we choose $R_{dedup} = 0.75$ and vary $R_{proto}$ (we discuss this choice in Section~\ref{sec:experimental_setup_details}). In Section~\ref{sec:results}, we compare the performance of D4 to baseline training and other methods, and in Section~\ref{sec:analysis} we analyze D4 and show that reclustering after semantic de-duplication indeed reduces the impact of duplicate-driven clusters (see Figure~\ref{fig:cluster_vs_no_recluster_d4}).

%% file: 3_results.tex
\section{Results}
\label{sec:results}

\begin{figure}[h]
\begin{center}
\includegraphics[width = 1.0\textwidth]{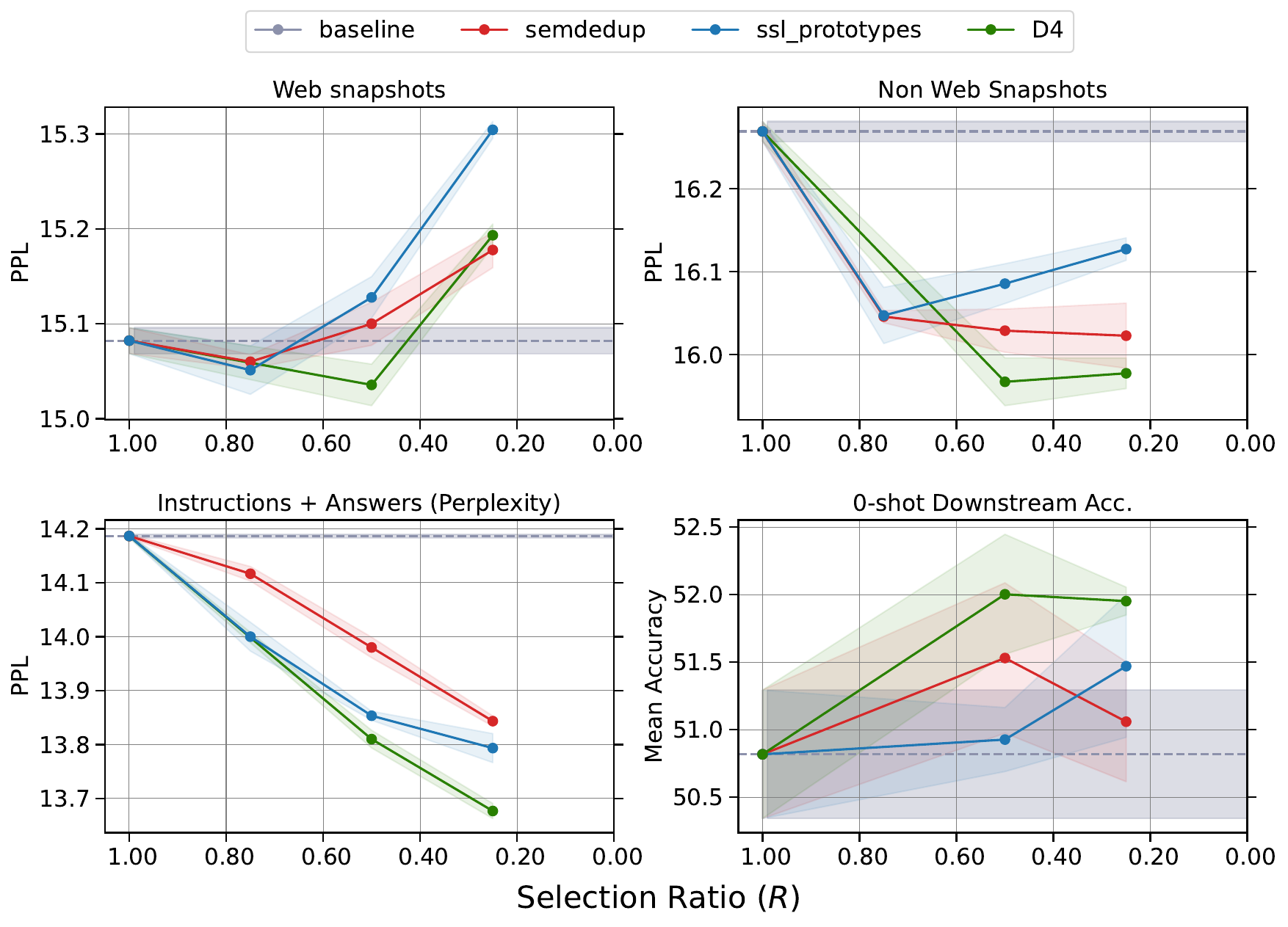}

\caption{Comparison of data selection methods on validation perplexity. Each point denotes a 1.3B OPT model trained on 40B tokens. The x-axis denotes the selection ratio $R$. The y-axis for the top 2 and bottom left graph depicts perplexity; the bottom right graph is average downstream on 16 NLP tasks from \citet{Zhang2022OPTOP}. The grey line denotes the value for baseline training. Shaded error is standard error across 3 seeds. \textbf{Each point on this graph is trained on the same token budget}: when we decrease $R$, we jointly increase the size of the source dataset (e.g. choosing 1/4 of documents from a 4x'ed sized source dataset).}
\label{fig:comparison_data_selection_methods_1_3_b_40B}
\end{center}
\vspace{-1em}
\end{figure}

\subsection{Fixed compute regime: can data selection help on fixed token budgets?}
\label{sec:results_fixed_compute}
In this section, we consider the fixed compute setting, where we curate and train on a fixed token budget by jointly increasing the size of the source dataset $D_{source}$ and decreasing $R$ (the fraction of the $D_{source}$ which is selected), such that the target token budget remains constant. This setting is analogous to the most common paradigm for LLM training. As $D_{source}$ grows and $R$ decreases, we select from larger and larger initial datasets, resulting in a larger set of high-quality data points to select from and increasing the overall quality of the selected set. For clarity, we plot performance as a function of the ratio of the $D_{source}$ to $D_{target}$. For each setting, we evaluate the performance of a baseline, SemDeDup alone, SSL Prototypes alone, and our proposed method D4.

\textbf{Validation Perplexity.} In Figure~\ref{fig:comparison_data_selection_methods_1_3_b_40B}, we show that a relatively small amount of data selection using any of the three methods (small $R$) brings consistent improvements on all validation sets. However, as we increase $R$, we observe \textit{opposing effects} on web snapshot and non-web-snapshots validation sets. 
We analyze this discrepancy in-depth in Section~\ref{sec:analysis}. However, on the Instruct OPT validation set, which corresponds much more closely to the the high-quality generations we want our LLMs to achieve, we found that all three methods led to consistent and clear perplexity improvements. Notably, we found that while all three methods provided benefits, D4 outperformed using both SemDeDup and SSL Prototypes independently, with the most notable gains exhibited when the source dataset is around 4x the target dataset size. Given that D4 consistently improves with source dataset size, we estimate this gap to grow with source dataset size. 

\textbf{Downstream Task Accuracy.} In Figure~\ref{fig:comparison_data_selection_methods_1_3_b_40B}, we also report 0-shot downstream accuracy averaged across a suite of NLP tasks. While the high variance of downstream accuracy makes it challenging to identify clear trends in the performance of various models, we again observe that 0-shot downstream accuracy generally increases with source dataset size.

Our findings also hold at larger model scales. We pick our best-performing configuration from 1.3B OPT experiments (e.g., $R = 0.25$) and train 6.7B OPT models on 100B tokens. Figure~\ref{fig:6.7b_panel_learning_curve} shows the positive effects of applying D4 with $R = 0.25$ for a 6.7B model. The model trained on the pruned data reaches the same perplexity as the baseline model using 20\% fewer update steps on average and achieves a 2\% improvement in accuracy on our suite of downstream tasks at the end of the training - about as much difference as was reported by \citet{Zhang2022OPTOP} between the OPT and GPT-3 family of models on the same set of tasks (See Figure 3 of \citet{Zhang2022OPTOP}).

\subsection{Fixed data regime: what happens when we run out of data?}
\label{sec:results_repeated_data}

\begin{figure}[h]
\begin{center}
\includegraphics[width = 1.0\textwidth]{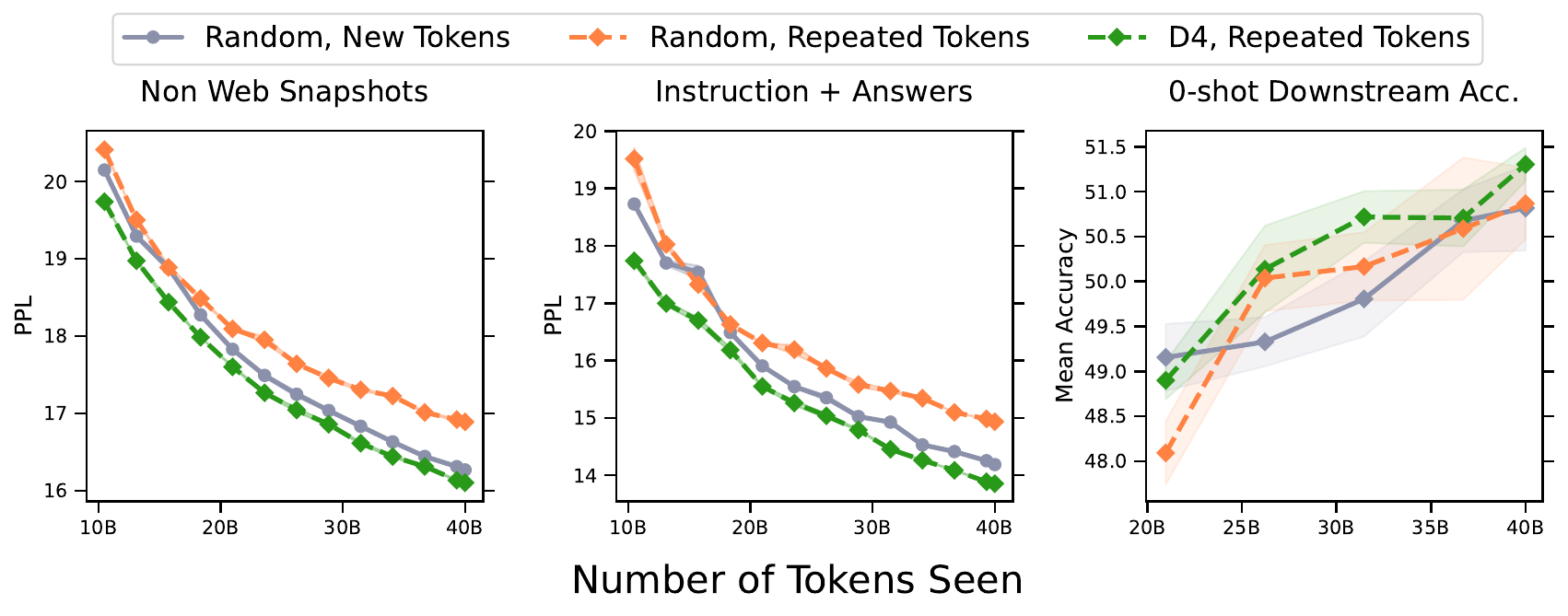}

\caption{Comparing new tokens vs. repeated tokens for random data selection and D4 for fixed selection ratio $R = 0.25$ for 1.3B OPT pre-training. Each method chooses 25\% of documents from the source dataset $D_{source}$, and epochs over that subset until the target token budget of 40B is reached. We observe that repeating tokens via D4 outperforms baseline training (random, new tokens).}
\label{fig:repeated_data_1.3b}
\end{center}
\end{figure}

The results in Section ~\ref{sec:results_fixed_compute}
indicate that, given a fixed amount of compute for training, selecting data from larger and larger source datasets is a promising method to improve language model performance. However, there is a practical limit to how much data can be curated from the web and, therefore, a natural limit to the size of the source dataset. What happens when we run out of data? \citet{Hernandez2022ScalingLA} found and analyzed disproportionately adverse effects of repeated data points in the training data. Similarly, concurrently to our work \citet{Muennighoff2023ScalingDL} shows that test loss deteriorates when epoching over a random subset of C4 more than four times. In this section, we investigate how the use of D4 affects model performance in this limited data, multi-epoch setting.

\begin{table}[t]
\centering
\begin{small}
\begin{tabular}{llll|lll}
    %\toprule
    $S$  & $T_{total}$ & $T_{selected}$ & Epochs &  Non-Web Snapshot PPL & Instruction + Answers PPL   \\
    \midrule
    \multirow{2}{*}{Random}  & 40B & 40B & $1$ & $16.27 \pm 0.012$ & $14.19 \pm 0.003$ \\ 
                             & 40B & 20B & $2$  & $16.39 \pm 0.011$ \textcolor{deepcarmine}{(+$0.12$)} & $14.37 \pm 0.015$ \textcolor{deepcarmine}{(+$0.18$)} \\

    \midrule 
    \multirow{1}{*}{D4} & 40B & 20B & 2  & $\textbf{16.10} \pm 0.024$ \textcolor{dartmouthgreen}{(-$0.17$)} & $\textbf{13.85} \pm 0.016$ \textcolor{dartmouthgreen}{($-0.34$)} \\
                          % & 160B & 40B & 1 & 19.83 & 13.73 & 13.67 \\
    %\bottomrule
\end{tabular}

\caption{For fixed data selection method and source dataset size, we compare the effects of choosing new tokens or repeating token. All models are 1.3B OPT models trained on 40B tokens. $T_{selected}$ denotes the number of tokens selected from the source dataset. The top row denotes baseline training. Mean and standard error across 3 seeds are shown. \textbf{Surprisingly, cleverly choosing tokens to repeat via D4 outperforms randomly selecting new tokens.}}
\label{tab:repeated_data}
\end{small}
\vspace{-1.5em}

\end{table}

To test this, we assume a fixed token budget and a fixed data size which matches the token budget. We evaluate training on all the data as well as for two epochs on subsets of the data selected either randomly or using D4. We trained 1.3B parameter OPT models on these configurations and report average perplexity in Table \ref{tab:repeated_data}. Unsurprisingly, epoching over a randomly selected subset of the data instead of using all the available data once leads to a slight degradation in model perplexity. In contrast, repeating data selected by D4 leads to an improvement in perplexity and downstream accuracy over randomly sampling new tokens. In other words, it is beneficial to select data via D4 and epoch 2 times, instead of doing one-pass learning on all available data. As seen in Figure~\ref{fig:repeated_data_1.3b}, this finding generally holds across training as well. We refer to Section~\ref{sec:extra_repeated_data_results_appendix} for results across model scale and data selection ratio.

To the best of our knowledge, this is the first result to demonstrate the benefits of repeating data for LLM pre-training, over randomly sampling new tokens via a principled data selection technique. We argue that the optimal way of using large-scale web data to pre-train LLMs could be: strategically choose a significantly smaller but better-distributed subset of the data and epoch over it multiple times.

\subsection{Cost of data selection}
\label{sec:cost_of_data_selection}

In Section \ref{sec:results_fixed_compute}, we find that by training a 6.7B parameter model on data selected by D4, we reach the final perplexity of a baseline model using 20\% fewer model updates. In our particular setup, this translates to \textbf{saving approximately 4300 GPU hours} - we will refer to this as the \textit{naive} efficiency gain as it does not account for the the cost of computing the selection metric. 
\begin{wrapfigure}[18]{r}{0.45\textwidth}

\begin{center}
\includegraphics[width = 0.35\textwidth]{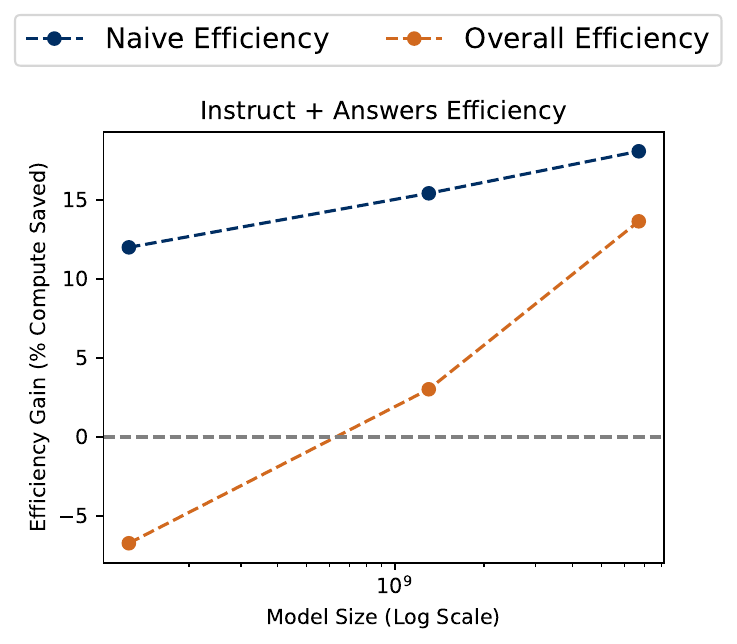}

\caption{\textit{Naive} and \textit{overall}  efficiency gain of data selection via D4 relative to the total cost of training as a function of model size on Instruct + Answers perplexity  at $R = 0.25$.}
\label{fig:efficiency_gain_vs_model_size}
\end{center}
\end{wrapfigure}

To demonstrate our method's practicality, we must ensure the cost of selecting data is significantly less than this. As described in Section~\ref{sec:methods}, selecting data via D4 involves: first, embedding documents via a 125M OPT model; second, computing K-Means indices + distance to indices. The first step is completed on a single machine with 96 CPU cores in approximately one day. Given the two orders of magnitude difference between the prices of CPU and GPU cores \footnote{Source: \url{https://aws.amazon.com/ec2/pricing/on-demand/}}, we consider this cost negligible. For the second step, embedding 400B tokens with a 125M parameter model takes approximately 888 GPU hours, using the same A100 GPUs. Subtracting this from the \textit{naive} efficiency gain of 4300 GPU hours, we arrive at an \textit{overall} efficiency gain of 3412 GPU hours. This is how much compute D4 saved us in practice when training our single 6.7B parameter model. In Figure~\ref{fig:efficiency_gain_vs_model_size}, we redo this calculation for different model sizes and we see that \textit{overall} efficiency gain increases with model size.  Based on this, we can conservatively estimate that D4 would have overall efficiency gains of 20\% for LLama-65B \citep{Touvron2023LLaMAOA} and 22\% for OPT-175B \citep{Zhang2022OPTOP}. 

\subsection{Analysis of D4}
\label{sec:analysis}

\subsubsection{Why does data selection hurt performance on web snapshots?}
\label{sec:web_analysis}

\begin{figure}[h]
\begin{center}
\includegraphics[width = 1.0\textwidth]{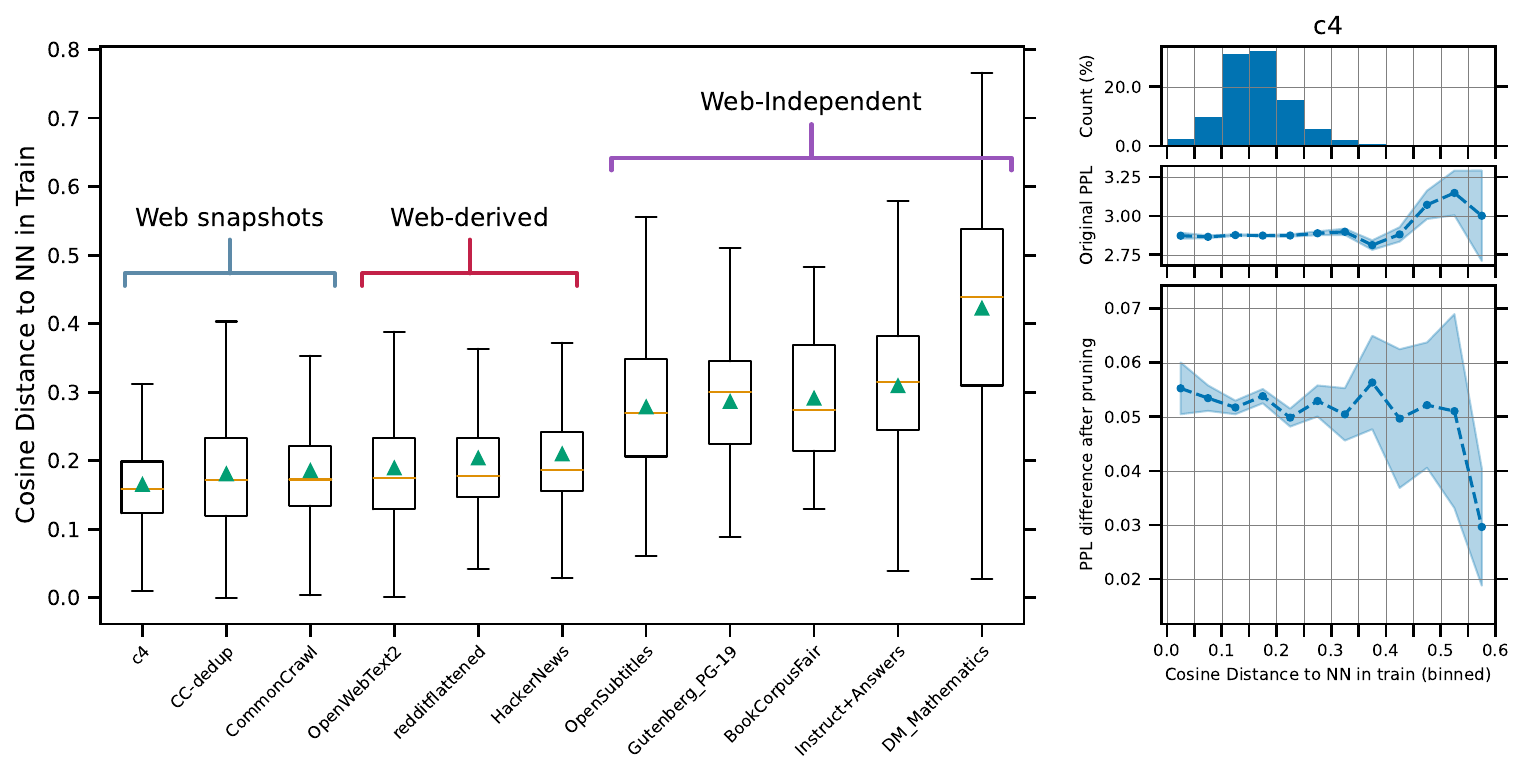}

\caption{\textbf{Left}: Train-test similarity across validation sets. X-axis denotes the name of the validation set (refer to Section~\ref{sec:methods} for more information about each validation set), and y-axis denotes the cosine distance to the nearest neighbor in the training set for the 1.3B OPT 40B baseline (the green triangle denotes mean, and the yellow bar denotes median). We observe that web-snapshots validation sets are closest to points in the training set. \textbf{Right}: Analysis of the C4 validation set. (Top): Histogram of cosine distance to nearest neighbor in train. For each bin, we show the mean original perplexity (middle) and mean difference in perplexity after data selection (bottom). "Easy" (low original ppl) points close to the training set are generally the points most affected by data selection.}
\vspace{-1em}
\label{fig:analysis_histogram_comparison_validation_sets}
\end{center}
\end{figure}

While we observe consistent \textit{average} perplexity improvements, Section~\ref{sec:appendix_individual_breakdowns_downstream_and_ppl} demonstrates that this perplexity improvement varies greatly across validation sets. More importantly, data selection always impairs performance on web snapshot validation sets such as CC-dedup, CommonCrawl, and C4. To investigate why this occurs, we embed each validation set into the same embedding space as the training set and search for the nearest neighbors to validation points in the training set for our 1.3B baseline model. In the left plot of Figure~\ref{fig:analysis_histogram_comparison_validation_sets}, we show that validation sets drawn from the same distribution as web-snapshots are closer to training set compared to other validation sets, while the right plot of Figure~\ref{fig:analysis_histogram_comparison_validation_sets} shows that data selection disproportionately affects these web-snapshot validation sets: on the top-right plot, we see that  web validation sets reside in regions of the embedding space which are sparsified as a result of data selection (e.g. regions of space close to cluster centroids in the training set), and in the bottom-right plot we see that these points are also the most affected by data selection, since their perplexity after data selection significantly increases. Moreover, the middle-right plot shows that these validation points have the lowest perplexity before pruning indicating that these points are "easy" points, perhaps due to their proximity to the training set.

\begin{figure}[h]
\begin{center}
\includegraphics[width = 0.33\textwidth]{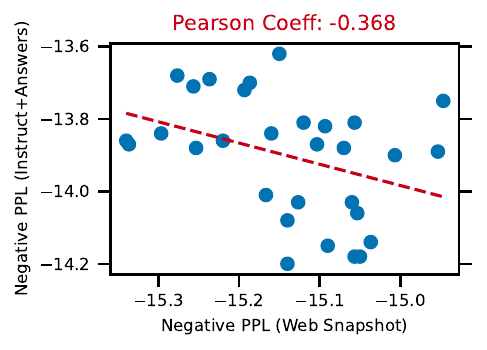}
\includegraphics[width = 0.32\textwidth]{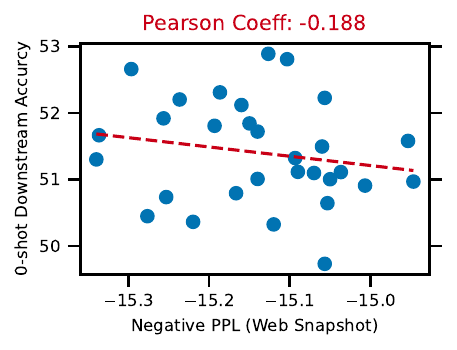}
\includegraphics[width = 0.33\textwidth]{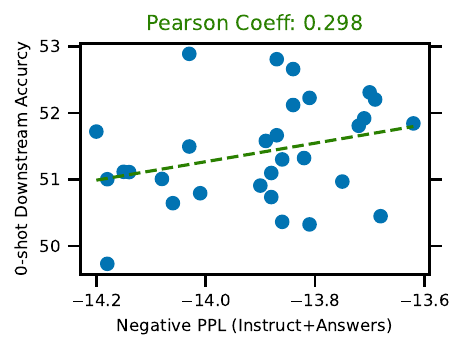}

\caption{Correlation between (left): negative Instruct+Answers perplexity and negative web snapshot perplexity, (middle): Downstream accuracy and negative web snapshot perplexity, (right): Downstream accuracy and negative Instruct+Answers perplexity. Each point is one training configuration (1.3B OPT model, 40B tokens), with the only change being the data selection method and pretraining seed. Web snapshot perplexity is slightly negatively correlated with stronger indicators of LM ability.}
\vspace{-1em}
\label{fig:scatterplot_web_snapshot_vs_prompts}
\end{center}
\end{figure}

Given that some of our validation sets are extremely close to the training set, we question whether they are still strong indicators of generalization. In fact, in Figure~\ref{fig:scatterplot_web_snapshot_vs_prompts}, we find evidence of a slight inverse relationship between perplexity on web snapshots and more robust indicators of LM ability, such as perplexity on instruction-tuned datasets and downstream accuracy. In contrast, we observe that perplexity on Instruct+Answers is positively correlated with downstream accuracy, suggesting that validation perplexity on instruction tuned data is a better measure of model quality. For this reason, we group most of our results in Section~\ref{sec:results} into Web Snapshots and Non-web Snapshots (which consists of  Web-Derived + Web-Independent from Figure~\ref{fig:analysis_histogram_comparison_validation_sets}, see Section~\ref{sec:appendix_explicit_validation_set_lists} for a full-list of validation set names).

\subsubsection{Importance of re-clustering between SemDeDup and SSL Prototypes}

\begin{figure}[h]
\begin{center}
\includegraphics[width = 1.0\textwidth]{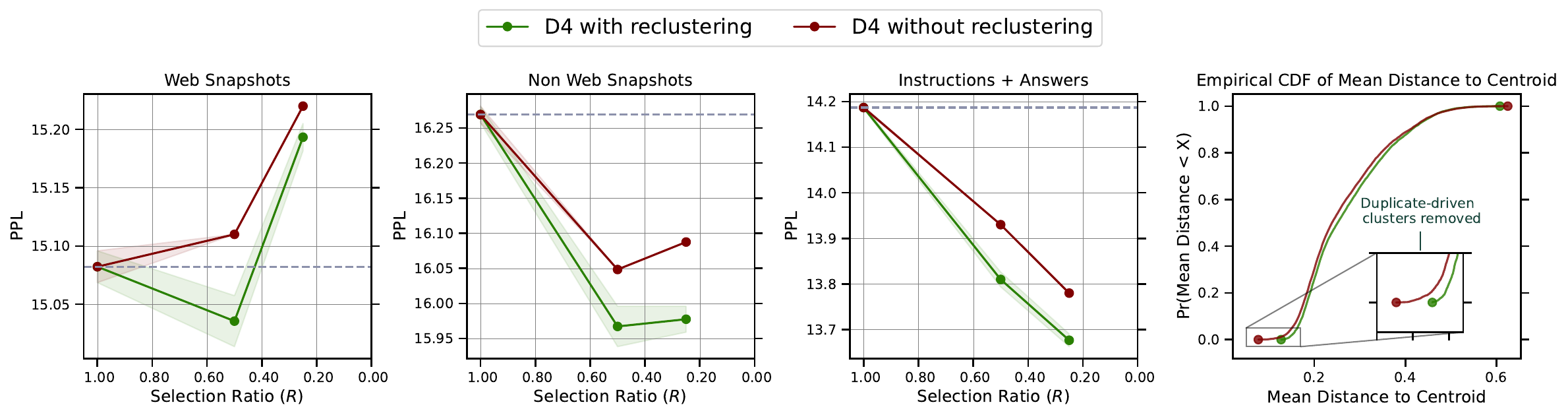}

\caption{Investigating the necessity of the re-clustering step in D4. We see that re-clustering improves perplexity across Web snapshots (left), Non-web snapshots (middle-left), and Instruct + Answers (middle-right). Right: Empirical CDF of mean distance to centroid, with and without re-clustering. Re-clustering removes duplicate driven clusters (clusters with low mean distance to centroid).}
\vspace{-1.5em}
\label{fig:cluster_vs_no_recluster_d4}
\end{center}
\end{figure}

As mentioned in Section~\ref{sec:methods}, we hypothesize that sparsifying dense regions of space containing excessive semantic duplicates improves the clustering quality and is, therefore, critical to the performance of D4. To isolate the effect of re-clustering on D4, we run experiments with a version of D4 where we remove the re-clustering step (e.g. we keep the original clustering). As shown in Figure~\ref{fig:cluster_vs_no_recluster_d4}, omitting the re-clustering step significantly worsens performance, and we observe in the rightmost plot of Figure~\ref{fig:cluster_vs_no_recluster_d4} that SemDeDup indeed removes extremely dense clusters surrounding centroids (e.g. duplicate-driven clusters). We analyze this in more depth in Section~\ref{sec:appendix_examples_duplicate_driven_clusters}.

%% file: 6_limitations.tex
\section{Summary and Limitations}
\label{sec:summary_and_limitations}

We introduced D4, a method for data curation on LLMs that improves training efficiency by ~20\% across multiple model scales, with larger gains at increased model scale. We also demonstrated that, in contrast to common practice, repeating data via epoching can be beneficial for LLM training, but only if the data subset is intelligently selected. 
While we have shown encouraging efficiency gains and performance improvements via D4, our work has several limitations and many future directions.

\textbf{Mixing different training distributions:} While we chose one data distribution to both select data and train on, modern LLM setups usually mix different data sources. Our method is likely complimentary to such pipelines: practitioners may use D4 to diversify and de-duplicate individual data sources and then mix data sources to provide additional diversity in their training dataset. We leave exploring the efficacy of D4 on a mix of training distributions as future work, but expect that this will yield further gains by reducing redundancy across datasets as well as within datasets.

\textbf{Model scale:} Due to compute limitations, the largest models we evaluated were 6.7B parameters trained on 100B tokens. While, to our knowledge, this is the largest to date application of embedding based data curation approaches, further investigation at model scales exceeding 100B would be very interesting, particularly in light of our observation that the efficiency gain grows with model scale.

%% file: appendix.tex
\renewcommand{\thefigure}{A\arabic{figure}}
\setcounter{figure}{0}
\renewcommand\thetable{A\arabic{table}}
\setcounter{table}{0}

\section{Appendix}

\subsection{Experimental Setup Details}
\label{sec:experimental_setup_details}

\subsubsection{Hyperparameters for model training}
As mentioned in Section~\ref{sec:methods}, we use the same hyperparameters and configurations as the original OPT model architecture from \citet{Zhang2022OPTOP}. We describe these hyperparameters briefly in Table~\ref{table:model_training_hyperparams}.
We chose these configurations because they are openly available and have been used as the standard in many previous works \citep{Abbas2023SemDeDupDL, dettmers2022llm, liu2023deja, tirumala2022memorization, Zhang2022OPTOP}. All models use GELU activation \citep{hendrycks2016gaussian}, Adam optimizer \citep{kingma2014adam} with $\beta_1 = 0.9$, $\beta_2 = 0.95$, $\epsilon = 10^{-8}$, weight decay set 
to $0.1$, and we clip gradient norms at 1.0. We use a polynomial learning rate schedule, where learning rate warms up from 0.0 to peak learning rate over the first 375 million tokens, and is then annealed to ($0.1$ * Peak LR) over the remaining $(T_{target} - 375)$ M tokens. We train all our models in fully sharded data parallel mode \citep{artetxe2021efficient} using Megatron-LM Tensor Parallelism \citep{shoeybi2019training} with fp16 precision. For reproducibility (and perhaps the only difference from the original configuration in \citet{Zhang2022OPTOP}) is that we do not use dropout.

\begin{table}[htbp]
\centering
  \caption{Model architecture details. Most of the parameter configurations are the same as in Table 1 of \citet{Zhang2022OPTOP}. Batch size denotes the total tokens that the model sees during one gradient descent update.}
  \label{table:model_training_hyperparams}
  \begin{tabular}{c c c c c c }
  \toprule
    \textbf{Scale} & Num Layers & Num Heads & Embedding Dim & Peak Learning Rate (LR) & Batch Size \\
    \midrule
    8M & 4 & 2 & 128 & 1.0e-3 & 0.5M \\
    125M & 12 & 12 & 768 & 6.0e-4 &  0.5M \\
    1.3B & 24 & 32 & 2048 & 2.0e-4 &  1M \\
    6.7B & 32 & 32 & 4096 & 1.2e-4 &  2M \\
  \end{tabular}
\end{table}

\subsubsection{Dataset Curation Details}

In this subsection, we describe how we curate \textit{CC-dedup}, the starting source dataset used throughout the paper. We start with 5 CommonCrawl dumps \footnote{https://commoncrawl.org/the-data/get-started/} which range from 2017 to 2020. We then use CC-net \citep{Wenzek2019CCNetEH}, to de-duplicate data at the paragraph level, remove non-English web pages, and filter out low-quality pages. The pipeline we use is identical to the pipeline used in \citet{Touvron2023LLaMAOA} (see the section after the subtitle "English CommonCrawl [67\%]", within Section 2).

On top of this, we add an additional step of MinHash \citep{broder1997resemblance} de-duplication at the document-level. The parameters for MinHash are 20 hashes per signature, 20 buckets, and 1 row per bucket. These parameters are the default parameters in the spark implementation of MinHashLSH, and we did not do a hyperparameter sweep on these parameters due to compute limitations. Previous work has attempted running MinHash with much more aggressive parameters: \citet{Lee2021DeduplicatingTD} and \citet{penedorefinedweb} use $20$ buckets, $450$ hashes per bucket, and $9000$ signatures per hash. We conjecture that more aggressive MinHash would remove more templates, resulting in a higher-quality starting dataset, potentially making the SemDeDup step of D4 less necessary. \citet{Abbas2023SemDeDupDL} did find that the performance of MinHash from \citet{Lee2021DeduplicatingTD} and SemDeDup are comparable at a fixed data selection ratio of 3.9\% on C4, indicating that SemDeDup filters out similar data to aggressive MinHash does. We leave sweeping over these hyperparameters as future work.

We note that since our dataset is curated from CommonCrawl dumps, there is risk that our training set contains offensive or PII content. We note, however, that this risk is no more than that of standard language modeling curation such as \citet{Touvron2023LLaMAOA}, since we use the same pipeline to filter CommonCrawl dumps.

\subsubsection{Parameters for Data Selection}
All methods introduced in Section~\ref{sec:methods} involve clustering embeddings using K-Means. Our starting training dataset CC-dedup contains roughly 600 million documents in total. Running K-Means clustering on all 600 million 768-sized vectors would take a considerable amount of compute. Instead, we follow previous work \citep{Abbas2023SemDeDupDL, Sorscher2022BeyondNS} and randomly sample roughly 100M documents with which to calculate centroids. We normalize the embeddings for these 100M documents to have L2-norm of 1.0, and then use faiss \citep{johnson2019billion} with the following parameters:

\begin{verbatim}
    faiss.Kmeans(
        768 # 125M OPT model embedding size, 
        11000 # 11K clusters, 
        niter=20 # 20 iterations, 
        verbose=True,
        seed=0, 
        gpu=False,
        spherical=True,
        min_points_per_centroid=1,
        max_points_per_centroid=100000000
    )
\end{verbatim}

\begin{figure}[h]
\begin{center}
\includegraphics[width = 1.0\textwidth]{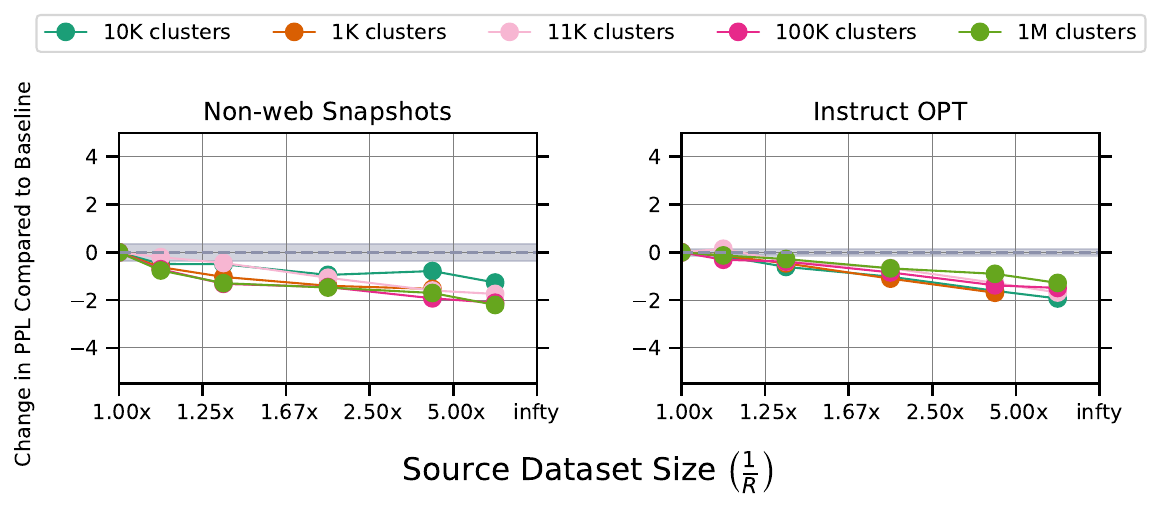}

\caption{Effect of number of clusters in K-Means on data selection performance. All models are 125M OPT models, where the training set (and starting source dataset) is C4 and we select data with SSL prototypes. The y-axis is the change in perplexity compared to baseline training, meaning that baseline training is at 0.0, and going \textit{down} on the graphs indicates \textit{better} performance. The x-axis is the source dataset size. We show results for average perplexity on Non-web snapshot validation sets (left) and Instruct + Answers (right). We notice that there is not a significant difference when changing number of clusters (e.g. if we drew error bars around each line, they would all be overlapping), but 11K clusters is generally among the top-3 best performing methods.}
\label{fig:n_clusters_sweep}
\end{center}
\end{figure}

We choose $11000$ clusters following previous work \citep{Abbas2023SemDeDupDL} and we note that this choice sticks to the heuristic that the number of clusters should roughly be the square root of the number of total points being clustered. We also note that in initial experiments for data selection at the 125M OPT model scale, we did not find a significant effect of number of clusters on the performance of our data selection methods (see Figure~\ref{fig:n_clusters_sweep}) this finding agrees with \citet{Abbas2023SemDeDupDL} who notice significant overlap between datasets selected by SemDeDup with different number of clusters (see Figure A2 in \citet{Abbas2023SemDeDupDL}).

We deliberately set min points per centroids low and max points per centroid high so that faiss does not attempt to manually balance the clusters while doing K-Means. \citet{Sorscher2022BeyondNS} found that explicitly class-balancing is important: they introduce the "class balance score" (see Section H of \citet{Sorscher2022BeyondNS}) which is the expectation of the quantity $\frac{\text{size of majority class}}{\text{size of minority class}}$ over all pairs of classes. They then set a hard limit for the class balance score of 0.5, meaning that "every class has at least 50\% of the images that it would have when pruning all classes equally" \citep{Sorscher2022BeyondNS}. We consider the unsupervised-learning analog of the class-balance score, which we refer to as the "cluster balance" score. The cluster balance score is the expectation of the quantity $\frac{\text{size of bigger cluster}}{\text{size of smaller cluster}}$ over all pairs of clusters. Across all of our data selection methods (and choices for R) we find that this value is generally equal to or bigger than $0.5$ without any explicit intervention. For this reason, we do not explicitly cluster balance, although we note that changing how many points are sampled from each cluster (based on properties of the cluster) is very interesting future work.

D4 parameters: The choice of parameters $R_{proto}$ and $R_{dedup}$ while using D4 will have impact on the performance of D4. Given limited compute, we are not able to sweep over these hyperparameters. Instead, we strategically choose these parameters: we first look at the highest value of $R$ in SemDeDup that results in perplexity improvement across validation sets. We choose the "highest value" because the purpose of SemDeDup is to remove duplicate-driven clusters and low $R$ with SemDeDup generally removes more than just templates/semantic duplicates. As seen in Section~\ref{sec:appendix_individual_breakdowns_downstream_and_ppl}, this generally occured with $R_{dedup} = 0.75$. Thus, we chose $R_{dedup} = 0.75$ and varied $R_{proto}$ to obtain different data selection ratios for D4.

\subsubsection{Which validation sets go into the averages?}
\label{sec:appendix_explicit_validation_set_lists}

For clarity, we explicitly state the validation sets which we consider "Web Snapshots", "Non Web Snapshots", and "Instruct + Answers" when reporting averages:

\textbf{Web Snapshots}: perplexity on validation set of C4, CC-dedup, CommonCrawl (from the Pile)

\textbf{Non-web Snapshots}: perplexity other validation sets from the Pile, comprising of OpenWebText2, HackerNews, Wikipedia (en), BookCorpusFair, DM Mathematics, Gutenberg PG-19, OpenSubtitles, and USPTO. Also included in this average is "redditflattened" (validation set from Pusshift.io Reddit \citep{baumgartner2020pushshift}), "stories", "prompts\_with\_answers" (which is described below) and "prompts" (which is the same as "prompts\_with\_answers" but where each sample is just the instruction-tuning prompt without the answer).

\textbf{Instruct + Answers}: perplexity on instruction-tuning data from OPT-IML \citep{Iyer2022OPTIMLSL}, where each sample contains both the instruction-tuning prompt and the answer (in Figure~\ref{fig:grid_1.3b_ppl} this is referred to as "prompts\_with\_answers." 

While the validation sets in web-snapshots and non-web snapshots are clear (they are either standard open-sourced datasets, or derived from commonly used data), we expect that the "Instruct + Answers" data might be new to some readers. We provide a few examples of what this validation set looks like in Table~\ref{fig:examples_from_instruct_opt}.

\begin{table}[H]
\begin{center}
\caption{Examples from "Instruct + Answers" validation set}
\label{fig:examples_from_instruct_opt}

\begin{tabular}{ p{14.0cm}  }
\toprule
\textbf{Raw Text}   
\\\midrule
Instructions: In this task, you are given two phrases: Head and Tail, separated with <sep>. The Head and the Tail events are short phrases possibly involving participants. The names of specific people have been replaced by generic words (e.g., PersonX, PersonY, PersonZ). PersonX is always the subject of the event. You have to determine whether the Head is located or can be found at/in/on the Tail or not. Classify your answers into "Yes" and "No". The phrase may also contain "\_\_\_", a placeholder that can be an object, a person, and/or an action.Input: Head: PersonX acknowledges gratefully the \_\_\_<sep>Tail: to use it Output: No
\\\hline Read the given sentence and if it is a general advice then indicate via "yes". Otherwise indicate via "no". advice is basically offering suggestions about the best course of action to someone. advice can come in a variety of forms, for example Direct advice and Indirect advice. (1) Direct advice: Using words (e.g., suggest, advice, recommend), verbs (e.g., can, could, should, may), or using questions (e.g., why don't you's, how about, have you thought about). (2) Indirect advice: contains hints from personal experiences with the intention for someone to do the same thing or statements that imply an action should (or should not) be taken. Input: Let it go. Output: yes"
\\\hline Instructions: You are given a sentence in English. Your job is to translate the English sentence into Italian. No! Demand to understand. Ask. Answer: No! Esigete di comprendere. Chiedete.
\\\hline Task: In this task you will be given a list of integers. You should round each integer to the nearest tens place. That means you should round the number to the nearest multiple of 10.Input: [528, -636, -686, 368, -433, 992, 886] Answer: [530, -640, -690, 370, -430, 990, 890]
\\
\bottomrule
\end{tabular}
\end{center}
\end{table}

\subsection{Efficiency gains across model scales and training}
\label{sec:efficiency_gains_across_model_scales_and_training}

\begin{figure}[t]
\begin{center}
\includegraphics[width = 1.0\textwidth]{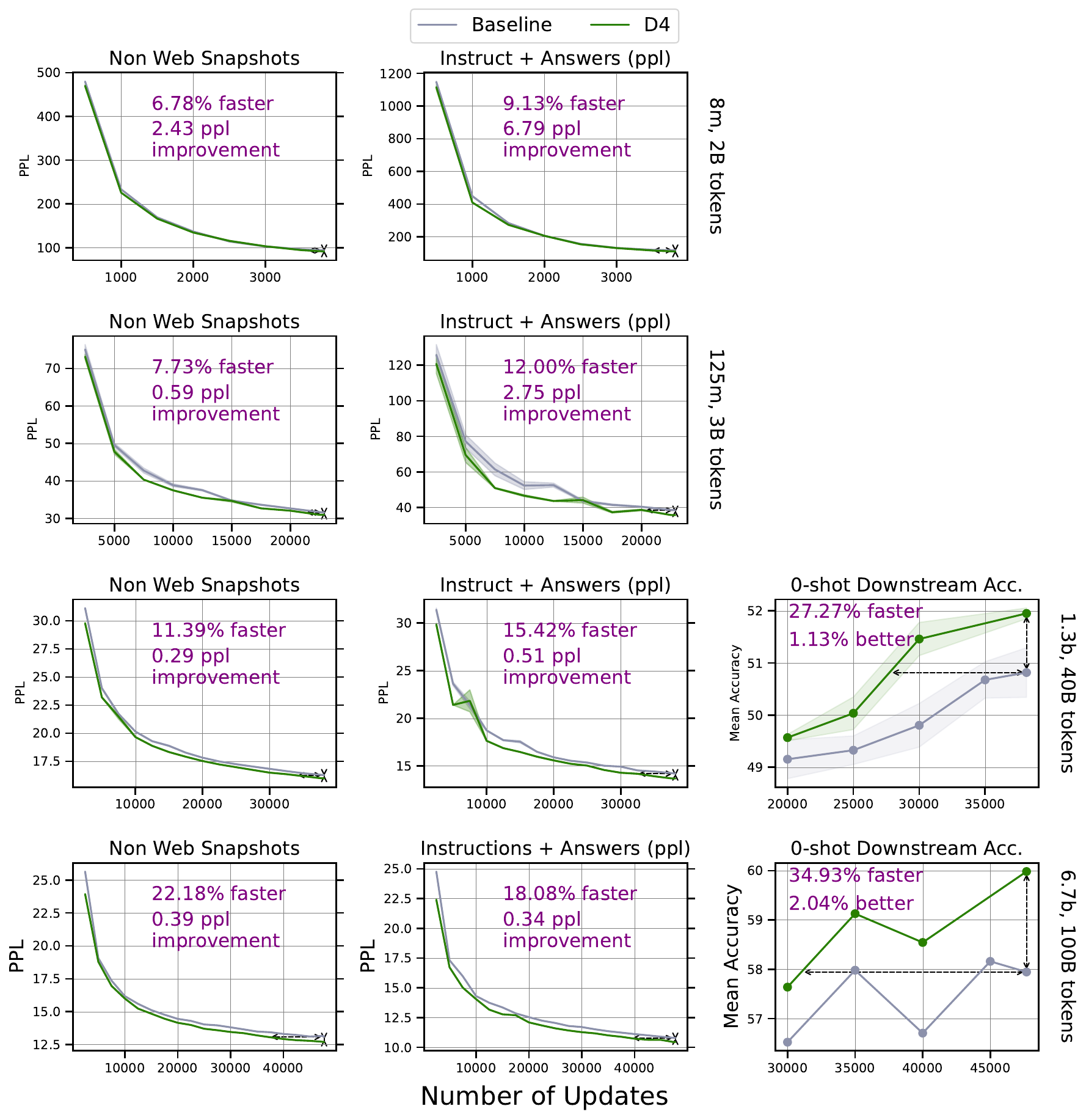}

\caption{Training trajectory of OPT models trained on raw data (gray line) and data selected via D4 (pink line). Across model scales (1st row: 8M OPT models trained on 2B tokens, 2nd row: 125M OPT models trained on 3B tokens, 3rd row: 1.3B OPT models trained on 40B tokens, 4th row: 6.7B OPT models trained on 100B tokens), we see significant efficiency gains in both perplexity (left two columns) and 0-shot downstream accuracy on 16 NLP tasks (right column). Importantly, we see that increasing model scale does not decrease efficiency gains. All plots show mean and standard error across three seeds, except for the last row. We do not evaluate downstream accuracy for models smaller than 1.3B because they are likely too close to random performance to indicate whether a particular data selection method is better. 
}
\label{fig:efficiency_across_model_scales_ppl}
\end{center}
\end{figure}

In this section, we investigate the relationship between model scale, and performance gain obtained by selecting data via D4. Specifically, we train three groups of models: 125M OPT models trained on $T_{target} = 3$B tokens, 1.3B OPT models trained on $T_{target} = 40$B tokens, and 6.7B OPT models trained on $T_{target} = 100$B tokens. We notice in Figure~\ref{fig:efficiency_across_model_scales_ppl} that D4 results in efficiency gains across the board in terms of perplexity. Surprisingly, these efficiency gains seem to increase with scale, indicating that at bigger model scales, D4 might lead to even more efficiency gains. We also see efficiency gains in 0-shot downstream accuracy for 1.3B and 6.7B model scales on the order of 30\% for both 1.3B and 6.7B models, but we note that evaluation downstream performance on intermediate checkpoints is not completely fair due to unfinished learning rate schedule. Nonetheless, we see that downstream accuracy efficiency gains are not decreasing with scale.

\newpage

\subsection{Individual Breakdowns of Downstream Accuracy and PPL} 
\label{sec:appendix_individual_breakdowns_downstream_and_ppl}

In Section~\ref{sec:results}, we see that D4, SSL prototypes, and SemDeDup achieves significant gains on perplexity (averaged across different validation sets) and downstream accuracy (averaged across different NLP tasks) compared to baseline training. Further, we generally see that D4 outperforms SSL prototypes and SemDeDup. In this section, we provide a more fine-grained analysis of these claims across individual tasks.

\begin{figure}[t]
\begin{center}
\includegraphics[width = 1.0\textwidth]{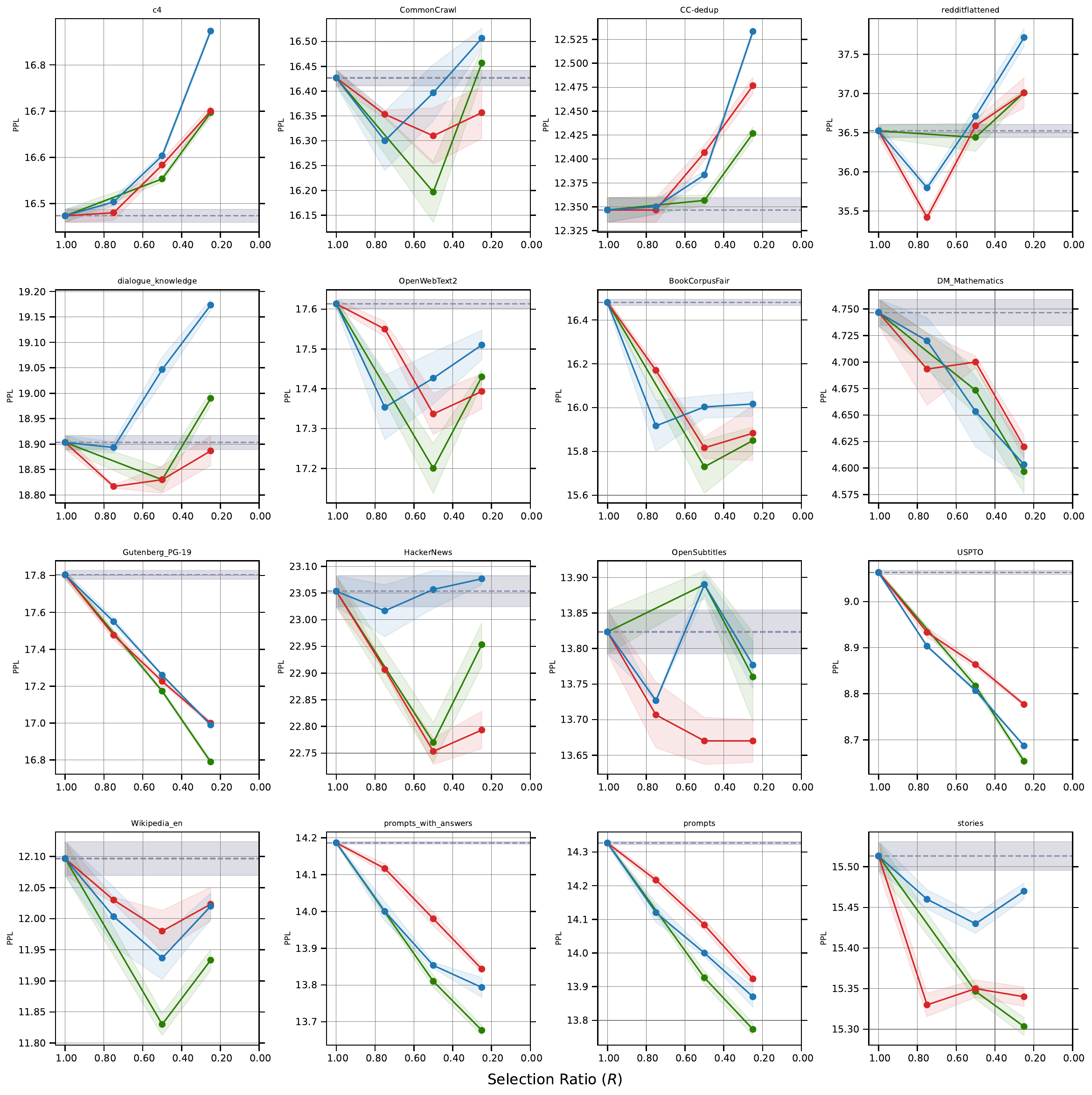}

\caption{Perplexity as a function of source dataset size for 1.3B OPT model 40B token training runs, across data selection runs. Each plot above represents perplexity on an individual validation set (see Section~\ref{sec:methods} for more information). Mean and standard error across 3 seeds is shown (standard error is denoted by shaded regions).}
\label{fig:grid_1.3b_ppl}
\end{center}
\end{figure}

For perplexity, we notice in Figure~\ref{fig:grid_1.3b_ppl}  that the claims in Section~\ref{sec:results} generally hold across validation sets: for web snapshots validation sets such C4, CC-dedup, and CommonCrawl, we see performance worsens with data selection compared to baseline training, and that D4 generally has the slowest rate of performance degradation. We note that, across all non web-snapshot validation sets, there is no clear winner among data selection methods. We emphasize however that \textit{we observe consistent improvement over baseline training on most validation sets} we use — for example in Figure~\ref{fig:grid_1.3b_ppl} we observe that, when selecting tokens from a 1.25x source dataset, all data selection methods improve over baseline across all validation sets except C4 and CC-dedup (however, as we explain in Section~\ref{sec:analysis}, this decrease in performance on C4 and CC-dedup is expected). 

\begin{figure}[t]
\begin{center}
\includegraphics[width = 1.0\textwidth]{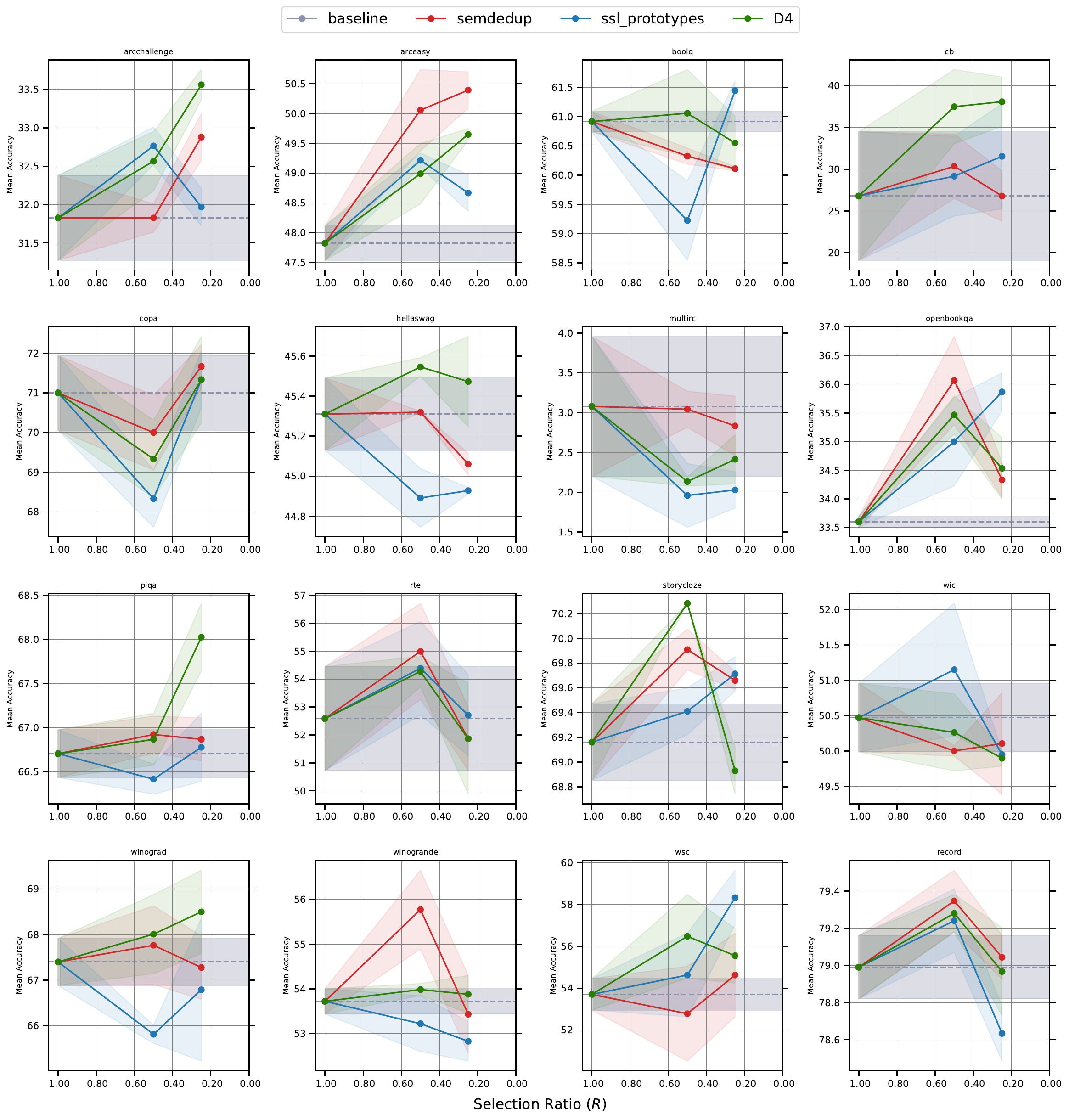}

\caption{Per-task breakdown of 0-shot downstream accuracy comparison across data selection methods, for 1.3B, 40B OPT model. For a description of the 16 NLP tasks shown above, see Section~\ref{sec:methods}. We note that there is considerable variability across individual downstream tasks.}
\label{fig:grid_1.3b_downstream_acc}
\end{center}
\end{figure}

For downstream accuracy, we chose to match the exact downstream evaluation done in \citet{Zhang2022OPTOP} since we use OPT architecture and hyperparameters. Similar to \citet{Zhang2022OPTOP}, we notice considerable variability across the 16 NLP tasks in Figure~\ref{fig:grid_1.3b_downstream_acc}, motivating us to look at the mean downstream accuracy across tasks.

\subsection{SSL prototypes and SemDeDup overlap}
\label{sec:appendix_analyzing_properties_of_clustering_raw}

\begin{figure}[h]

\begin{center}
\includegraphics[width = 1.0\textwidth]{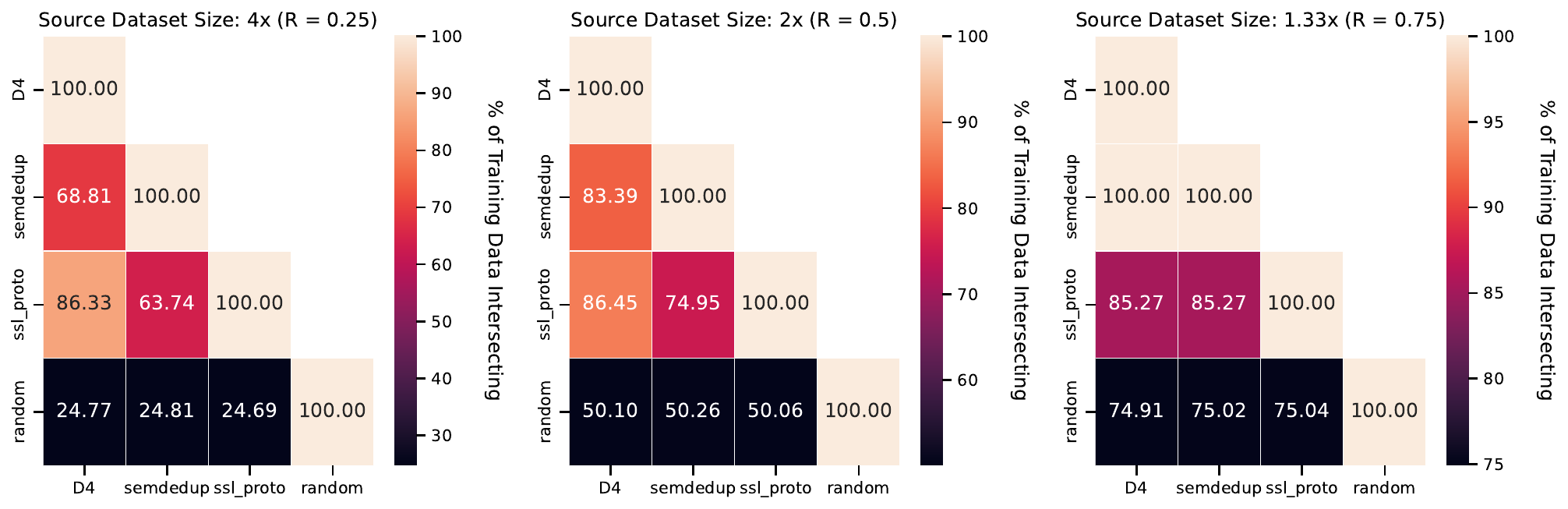}

\caption{Similarity between data selection methods. Each square represents the percentage of training data that is intersecting, when selecting data via two different strategies. The $x$ and $y$ axis enumerate different data selection strategies.}
\label{fig:correlation_plot}
\end{center}
\end{figure}

\begin{figure}[H]
\begin{center}
\includegraphics[width = 0.5\textwidth]{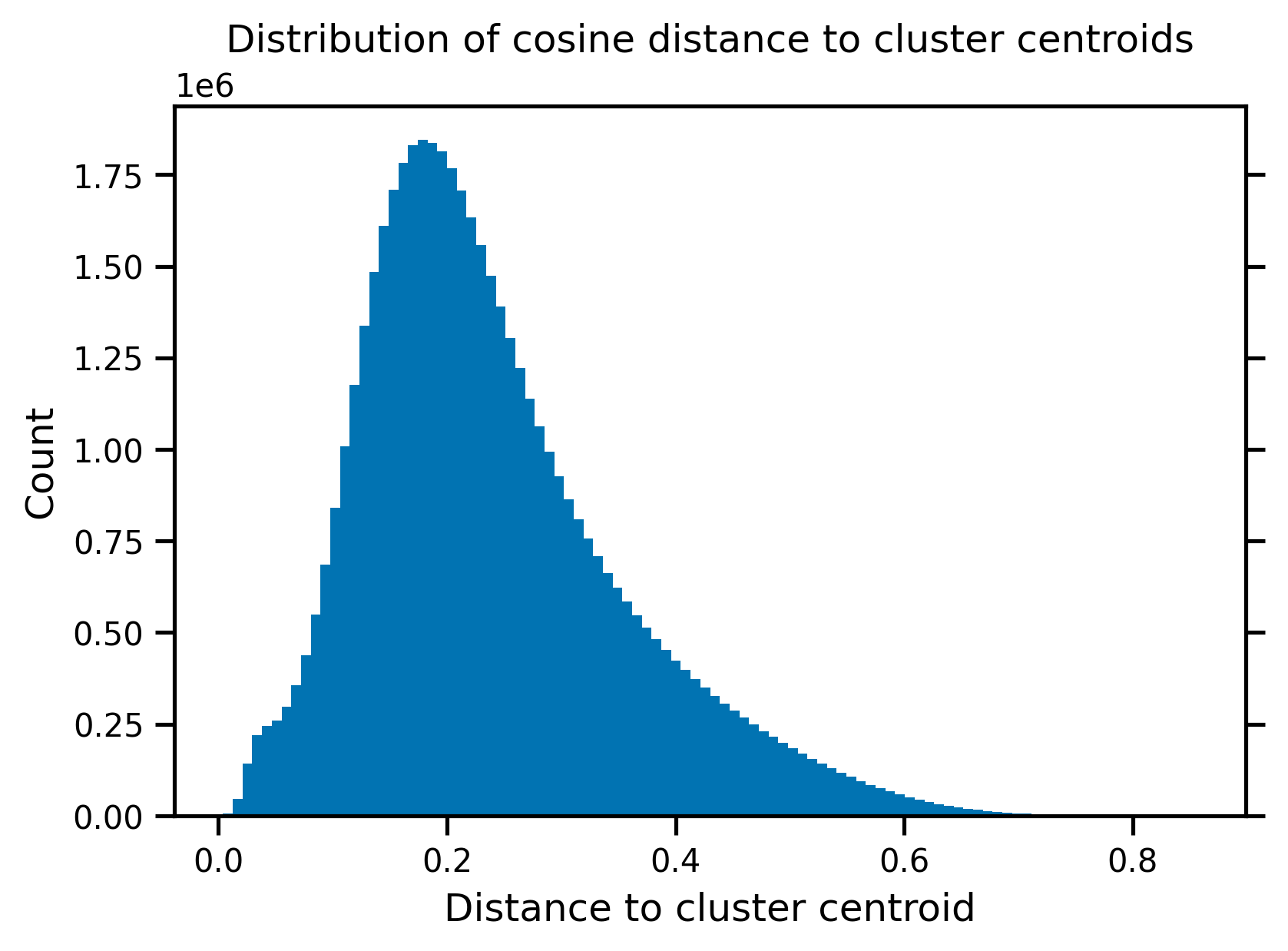}
\caption{Distribution of cosine distance to cluster centroids for ~50M randomly selected documents from the training set of CC-dedup. We notice that the distribution is skewed right, implying that datapoints are generally close to centroids.}
\label{fig:distribution_of_cluster_centroids}
\end{center}
\end{figure}

Figure~\ref{fig:correlation_plot} shows the overlap between datasets selected by SemDeDup and SSL Prototypes. While the two methods do not arrive at the same set of data points, there is a significant overlap between the datasets curated by the two methods. We hypothesize that this is because both SSL prototypes and SemDeDup prune away dense regions of space surrounding cluster centroids: by definition, SemDeDup sparsifies dense regions of space within a cluster; similarly, by definition, SSL prototypes will prune away datapoints close to the cluster centroids. Since K-means clustering places centroids in dense regions of space (see Figure~\ref{fig:distribution_of_cluster_centroids} where we observe that the distribution of cosine distances to cluster centroid is skewed right), we know that the regions of space surroundings centroids will be dense, and expect SSL prototypes and SemDedup to have significant overlap. Qualitatively, we inspect a few examples of points close to cluster centroids in Figure~\ref{fig:closest_to_centroid_682}, Figure~\ref{fig:closest_to_centroid_975}, Figure~\ref{fig:closest_to_centroid_10715}, and see that examples close to cluster centroids can be semantically redundant (e.g. templates). Therefore, it makes sense that any reasonable data selection strategy would prioritize sparsifying these dense regions of space surrounding cluster centroids. As mentioned in Section~\ref{sec:methods}, sparsifying these dense regions of space containing excessive semantic duplicates is the original motiviation behind D4. As shown in Figure~\ref{fig:cluster_vs_no_recluster_d4}, omitting the re-clustering step significantly worsens performance, and we observe in the rightmost plot of Figure~\ref{fig:cluster_vs_no_recluster_d4} that SemDeDup indeed removes duplicate-driven clusters.

\subsection{Investigating Train-Validation overlap}
\label{sec:investigating_train_validation_overlap}

\begin{figure}[H]
\begin{center}
\includegraphics[width = 1.0\textwidth]{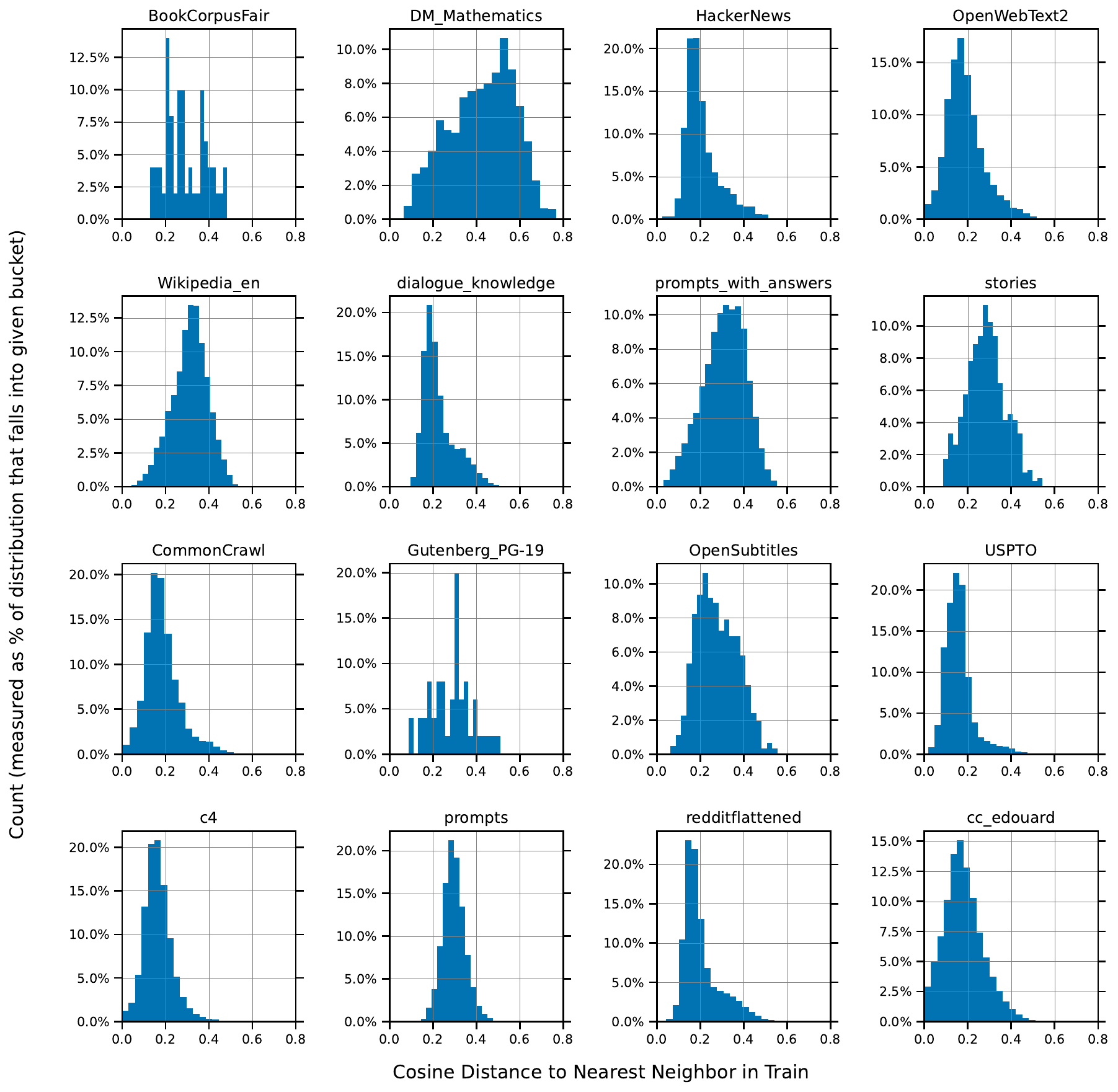}
\caption{Distribution of cosine distance to nearest neighbor in the training set, for each individual validation set.}
\label{fig:appendix_distribution_of_cluster_centroids_grid}
\end{center}
\end{figure}

As briefly described in Section~\ref{sec:analysis}, we observe that many of our validation sets are close (in cosine distance) to our training sets, and the impact of data selection is varies across individual validation sets. Individual validation sets live in different regions of the embedding space, and as such they are affected differently by data selection. For example, one could imagine that web-snapshot validation sets such as C4 is close to CC-dedup in the embedding space, while esoteric validation sets (such as Gutenberg PG 19 or DM Mathematics) might be far. To quantify this, we first find the nearest neighbors in the training set to each validation point in all of our validation sets. We then qualitatively check (see Table~\ref{fig:appendix_nn_to_valid_c4} and Table~\ref{fig:appendix_nn_to_valid_uspto} for examples) that nearest-neighbors in the training set truly convey information about validation points. we observe significant overlap between training points and validation points.  We then quanitatively analyze how close each validation set is to the training set: in Figure~\ref{fig:appendix_distribution_of_cluster_centroids_grid}, we show the breakdown of this distribution for each validation set. We see a general trend, that web-snapshots validation sets are closest to the training set as they are skewed to the right, while more esoteric validation sets (Gutenberg, or Wikipedia (en)) are more centered or even slightly left-skewed.

Motivated by this, we compare validation sets side-by-side (in terms of distance to training set) in Figure~\ref{fig:analysis_histogram_comparison_validation_sets}, and we see a similar trend. To further understand why different validation sets are affected differently by data selection, we loop through each data point in the validation set and record:
\begin{itemize}
    \item distance to the training set e.g. how close is the validation point to the training set
    \item perplexity difference before and after data selection with D4 e.g. how much was this validation point affected by data selection
    \item original perplexity e.g. how easy was this data point originally
\end{itemize}

\begin{figure}[H]

\begin{center}
\includegraphics[width = 1.0\textwidth]{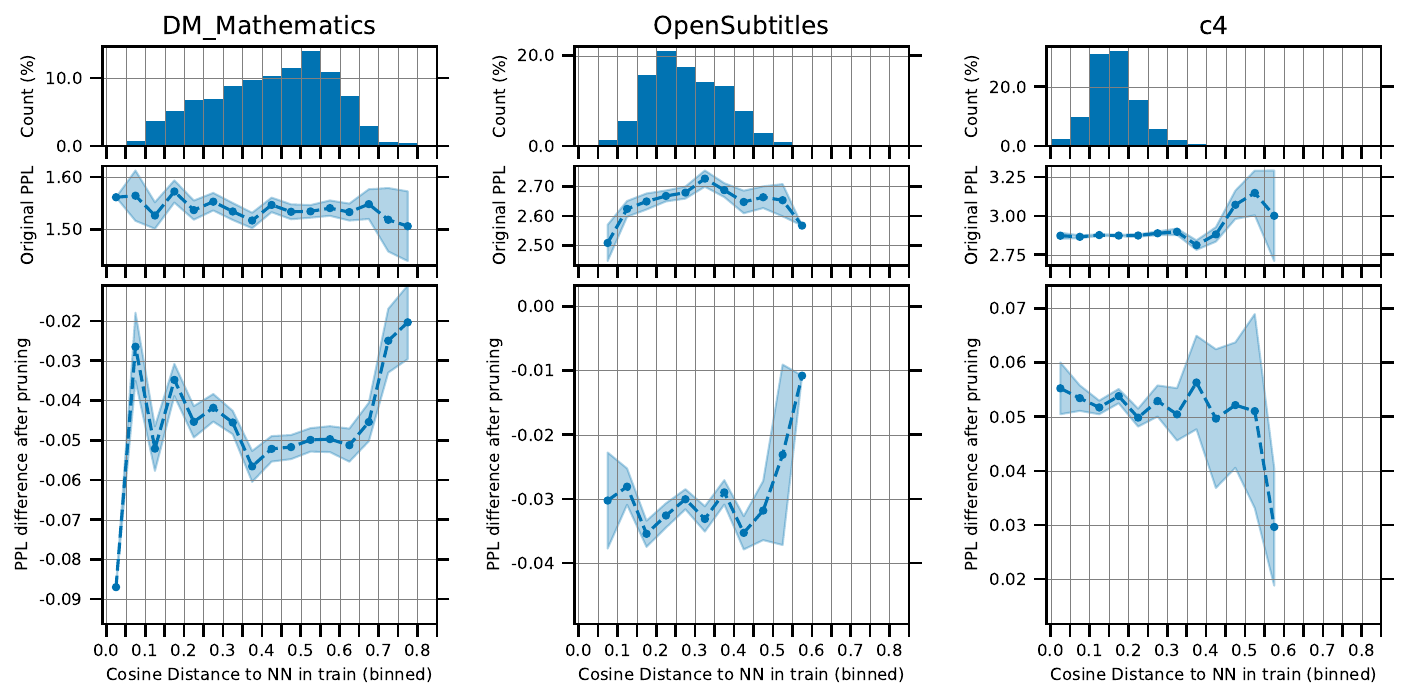}

\caption{(Top): Histogram of cosine distance to nearest neighbor in train. Within each bin, we show the mean original perplexity (middle) and mean difference in perplexity after data selection (bottom), for DM\_Mathematics (left), OpenSubtitles(middle), and C4 (right). We note that points in the C4 validation set closest to the training set are both "easy" (perhaps because of proximity to training points) and are affected the most by data selection. We do not see this trend for non-web snapshot validation sets such as DM\_Mathematics and OpenSubtitles.}
\label{fig:appendix_memoriztion_validation_sets}
\end{center}
\end{figure}

In Figure~\ref{fig:appendix_memoriztion_validation_sets}, we observe an interesting trend: for web-snapshot validation sets such as C4, the validation points closest to the training set are both (1) the easiest (lowest perplexity) points before data selection and (2) the points most affected by data selection. This seems to indicate that these validation points are "easy" due to their proximity to training points, and when these training points are removed from the training set due to data selection, the close-by validation points become difficult for the model. We do not see this trend on non-web snapshot validation sets such as DM Mathematics and Open Subtitles; in fact, we see an opposite trend where points furthest from the training set are generally most affected by data selection.

\begin{figure}[H]

\begin{center}
\includegraphics[width = 1.0\textwidth]{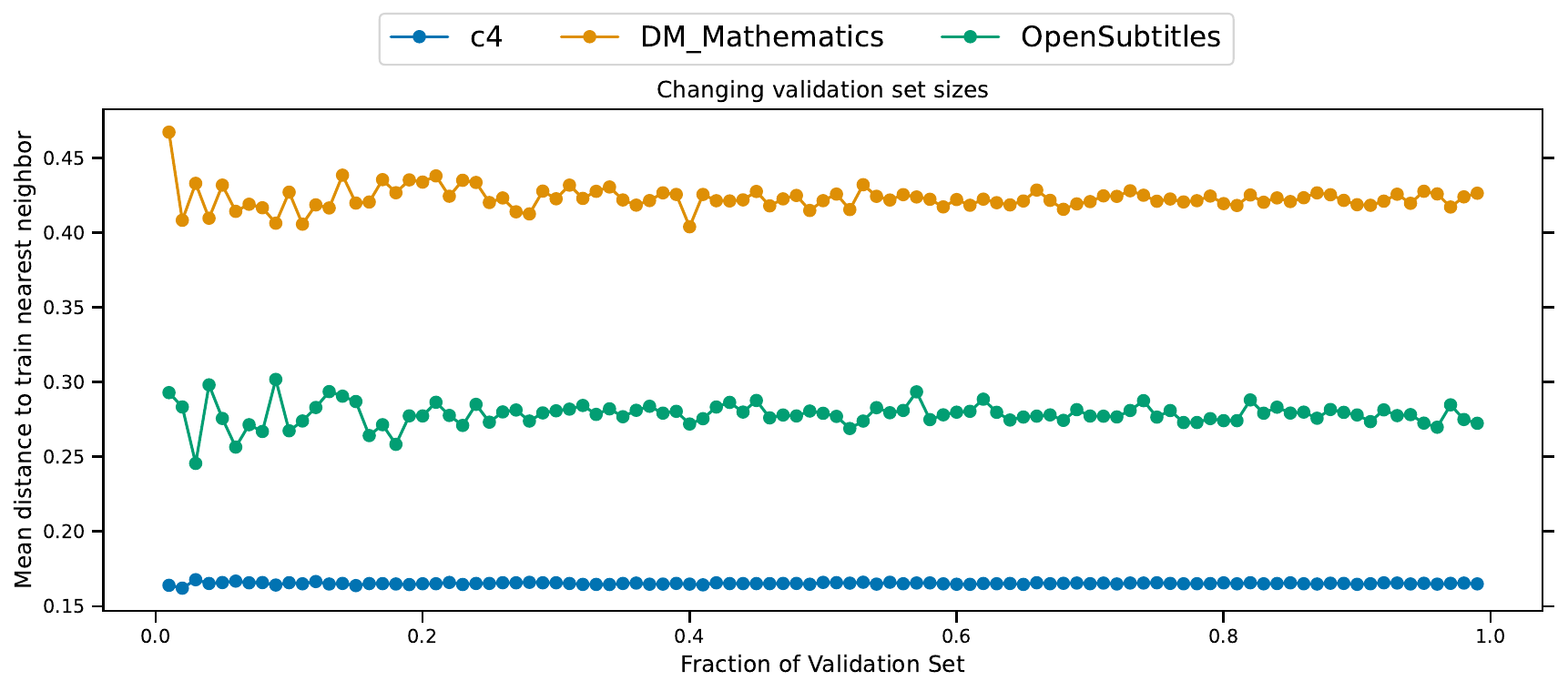}

\caption{Studying the effect of validation set size on cosine distance to nearest-neighbor in training set. On the x-axis, we vary the size of the validation set (by randomly sampling the original larger validation set), and the y-axis represents distance to nearest neighbor in the training set (averaged across the validation set). We observe that regardless of what fraction of the original validation set is sampled, the mean distance to the nearest neighbor in train does not change, indicating that Figure~\ref{fig:analysis_histogram_comparison_validation_sets} is not due to different validation set sizes.}
\label{fig:rebuttal_effect_validation_set_size}
\end{center}
\end{figure}

\begin{figure}[H]

\begin{center}
\includegraphics[width = 1.0\textwidth]{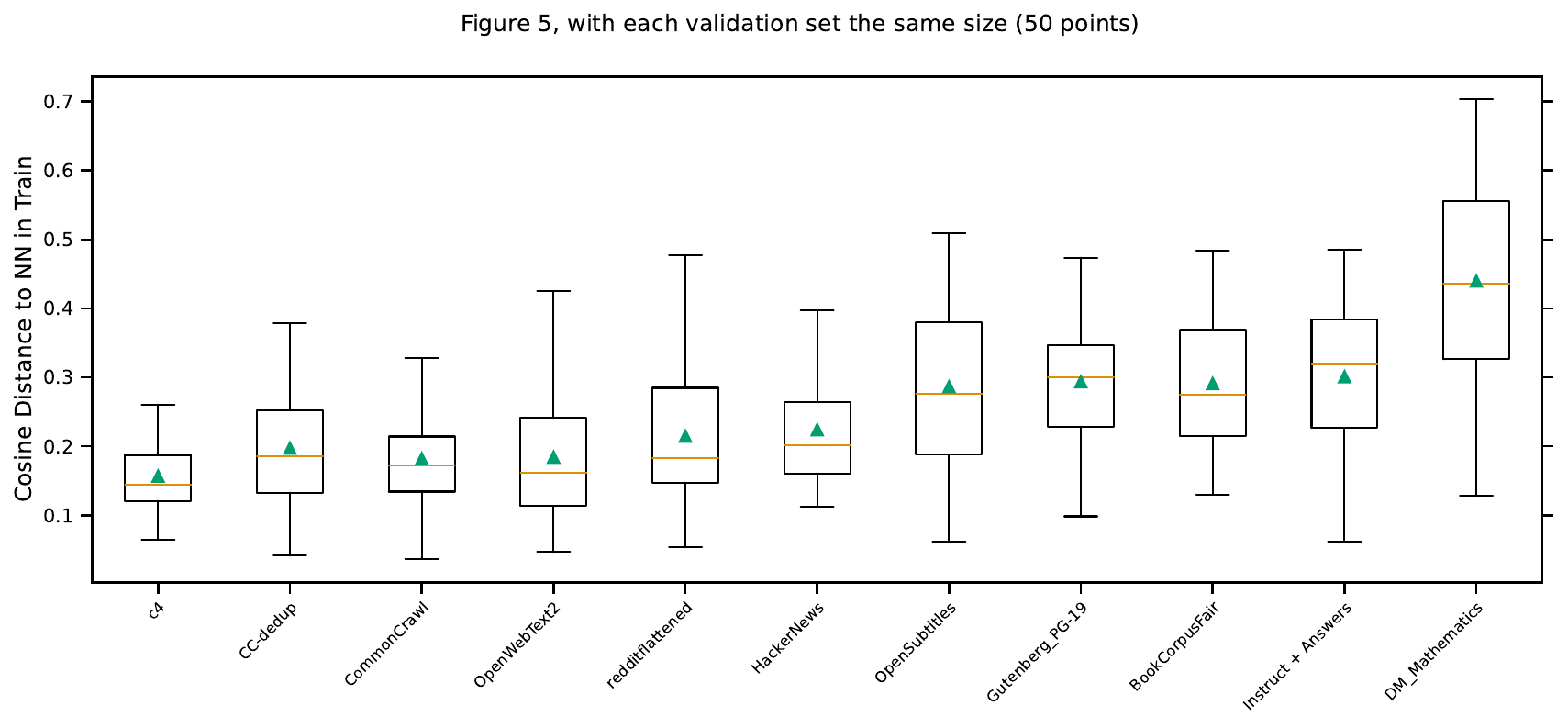}

\caption{Investigating whether Figure~\ref{fig:analysis_histogram_comparison_validation_sets} changes if we control for validation set size. In the Figure above, each validation set contains 50 data points, which is the size of the smallest validation set we use (BookCorpusFair). If a validation set is bigger than 50 data points, we randomly sample the validation set to obtain 50 data points.}
\label{fig:rebuttal_redo_figure_analysis_same_valid_size}
\end{center}
\end{figure}

As a sanity check, we change the sizes of validation sets used to plot Figure~\ref{fig:analysis_histogram_comparison_validation_sets} in Section~\ref{sec:analysis}. We see in Figure~\ref{fig:rebuttal_redo_figure_analysis_same_valid_size} that controlling for validation set size, we get the same jump going from web-derived to web-independent validation sets. In running this experiment, we are forced to randomly sample if the particular validation set is too big; to ensure that such random sampling does not change the distance to nearest neighbor in the training dataset too much, we vary the amount we sample for three differently sized datasets in Figure~\ref{fig:rebuttal_effect_validation_set_size}. We observe that changing the amount we randomly sample from a validation set does not significantly change the mean distance to nearest neighbor in train.

\begin{figure}[H]

\begin{center}
\includegraphics[width = 1.0\textwidth]{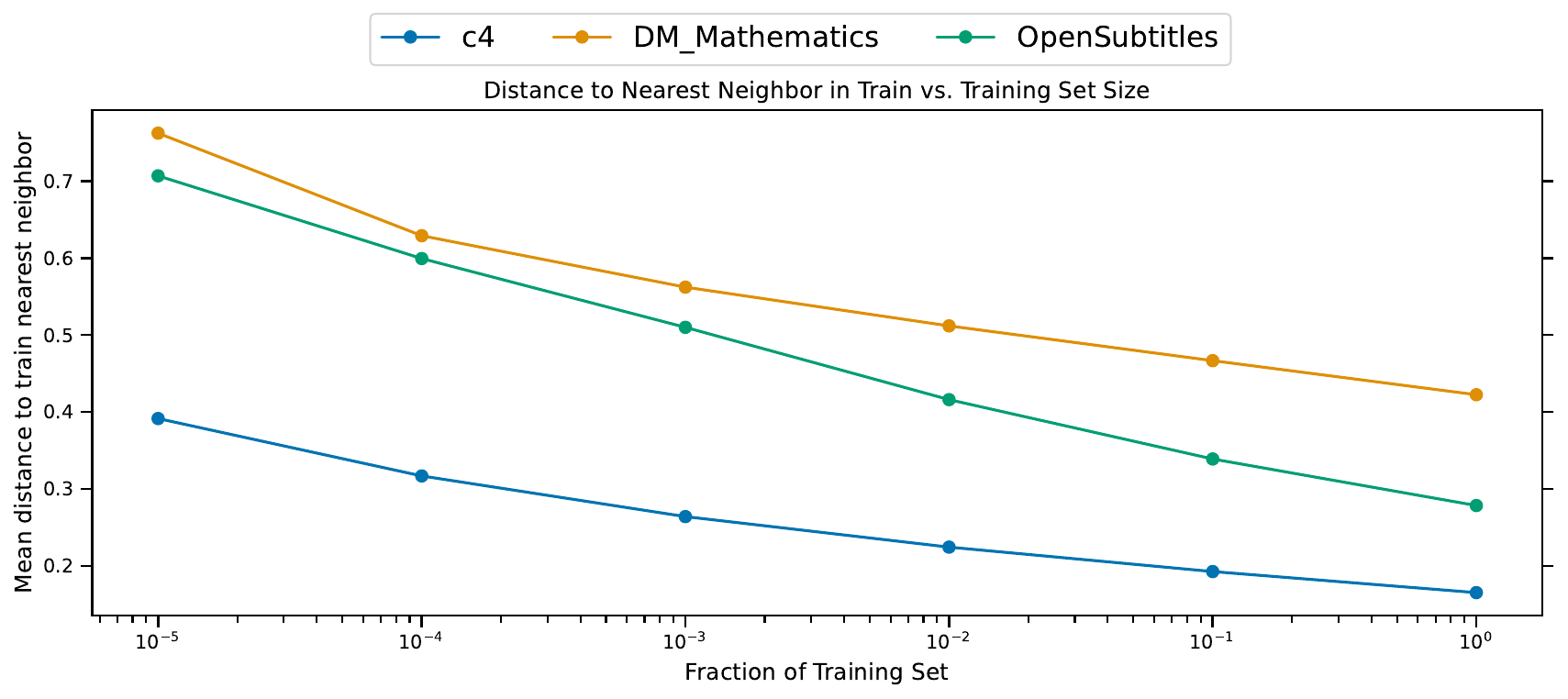}

\caption{Studying the effect of training set set size on cosine distance to nearest-neighbor in training set. On the x-axis, we vary the size of the training set (by randomly sampling the original training set), and the y-axis represents distance to nearest neighbor in the training set (averaged across the validation set). We observe that cosine distance to the training set increases with smaller training sets, but the relative ordering of validation sets (with respect to mean distance to training set) remains the same.}
\label{fig:rebuttal_effect_train_set_size_a}
\end{center}
\end{figure}

\begin{figure}[H]

\begin{center}
\includegraphics[width = 1.0\textwidth]{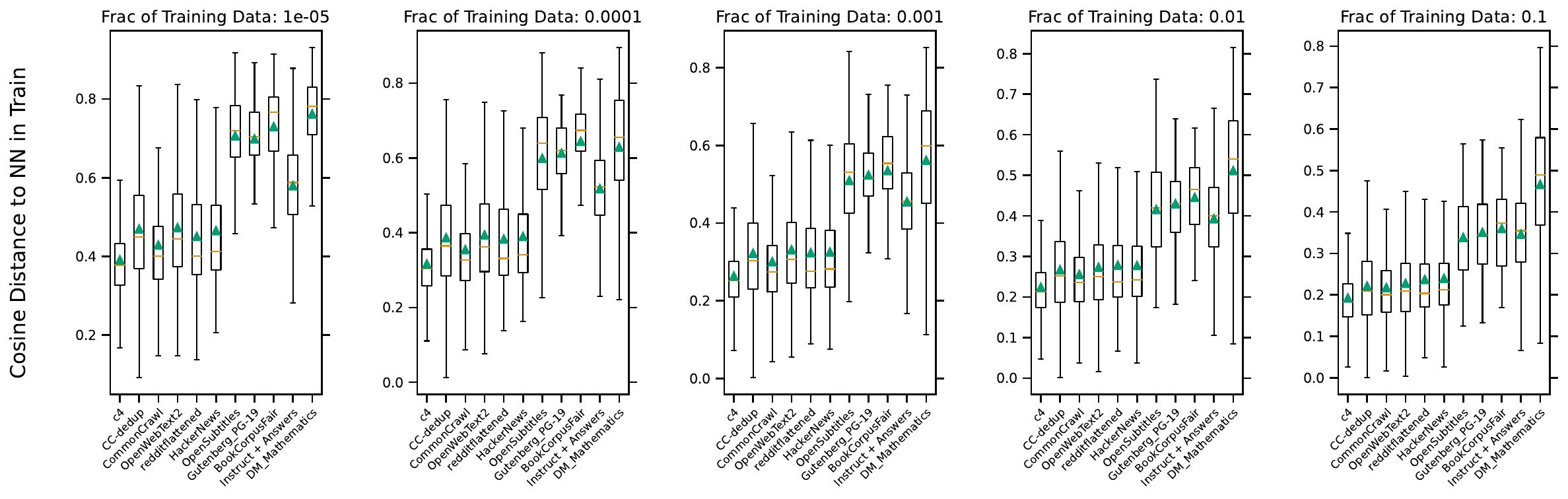}

\caption{Investigating whether Figure~\ref{fig:analysis_histogram_comparison_validation_sets} changes if we change training set size set size. In the figure above, each plot randomly samples a fraction of the training set (the fraction is denoted by the title of the plot). We see that the relative ranking of the validation sets generally remains the same, and there is consistently a jump between web-derived and web-independent validation sets.}
\label{fig:rebuttal_effect_train_set_size_b}
\end{center}
\end{figure}

We also investigate whether the differences between validation sets in Figure~\ref{fig:analysis_histogram_comparison_validation_sets} is due to training set size. We would expect that smaller training sets are "further" from validation sets, since (). Indeed we see this in Figure~\ref{fig:rebuttal_effect_train_set_size_a}. However, we observe that the relative ordering of validation sets (with respect to average distance to the training set) remains the same for any fixed training dataset size. Moreover, we see in Figure~\ref{fig:rebuttal_effect_train_set_size_b} that the relative ranking of all validation sets as well as the jump from web-derived to web-independent validation sets from the original Figure~\ref{fig:analysis_histogram_comparison_validation_sets} holds, even as we reduce training dataset size.

\subsection{Further investigation of repeating tokens}
\label{sec:extra_repeated_data_results_appendix}

In this section, we investigate whether the findings from Section~\ref{sec:results_repeated_data} hold across model scale, data selection ratio (e.g. number of epochs), and data selection method.

\begin{figure}[h]
\begin{center}
\includegraphics[width = 1.0\textwidth]{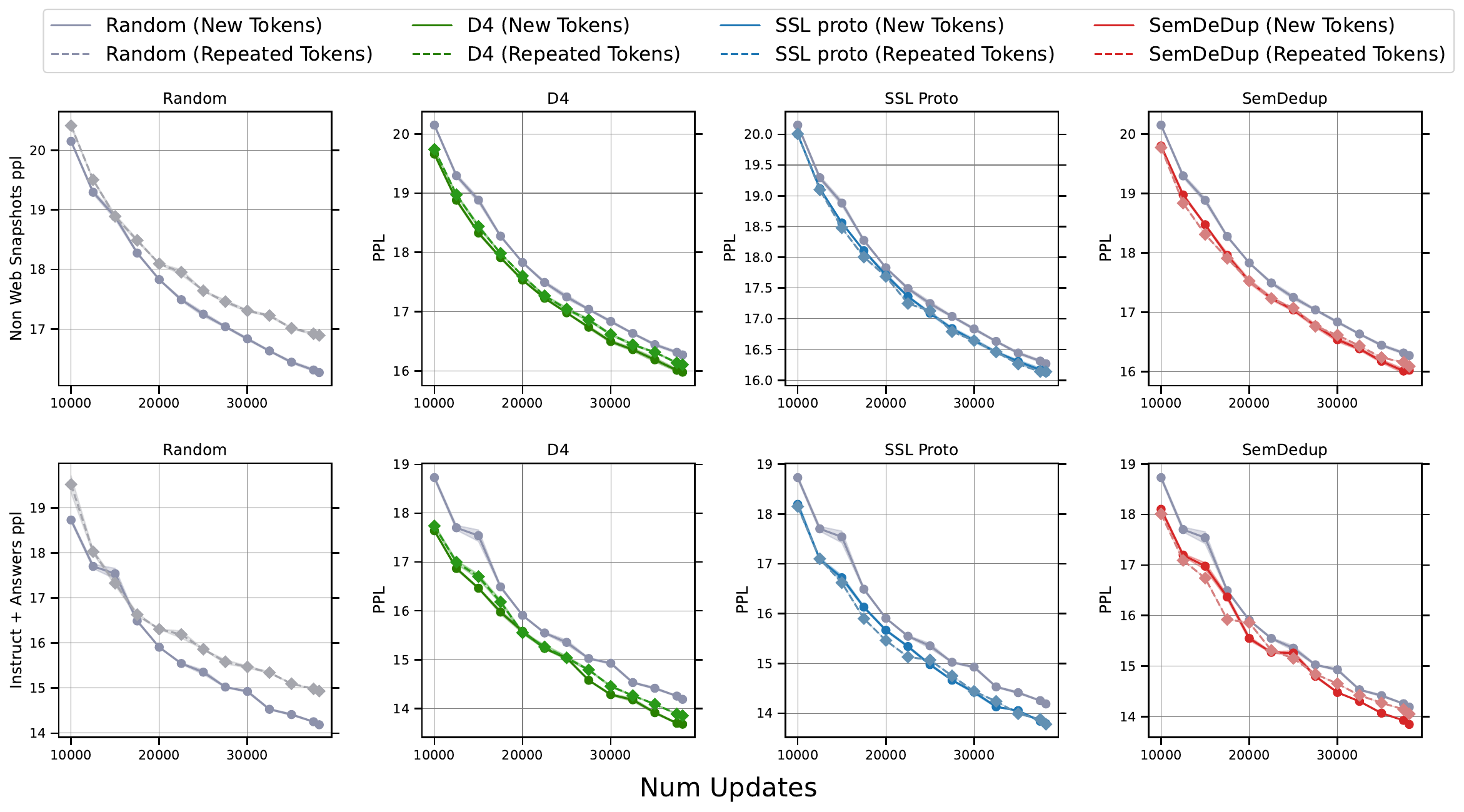}

\caption{Effect of repeating tokens across data selection methods over training. X-axis denotes the number of updates, and the y-axis denotes average perplexity across non-web-snapshot validation sets (top row) and Instruct OPT (bottom row). Each column in the plot above denotes a different data selection method. Within each column: (1) the gray line denotes baseline training, (2) the colored-dashed line denotes repeating tokens via the specified data selection method, and (3) the colored-solid line denotes selecting new tokens via the specified data selection method. Repeating data is generally worse than selecting new data for a \textit{fixed data selection method} (e.g., fixed column).}
\label{fig:appendix_repeated_data_new_tokens_always_better_non_web_snapshots}
\end{center}
\end{figure}

\begin{figure}[h]
\begin{center}
\includegraphics[width = 1.0\textwidth]{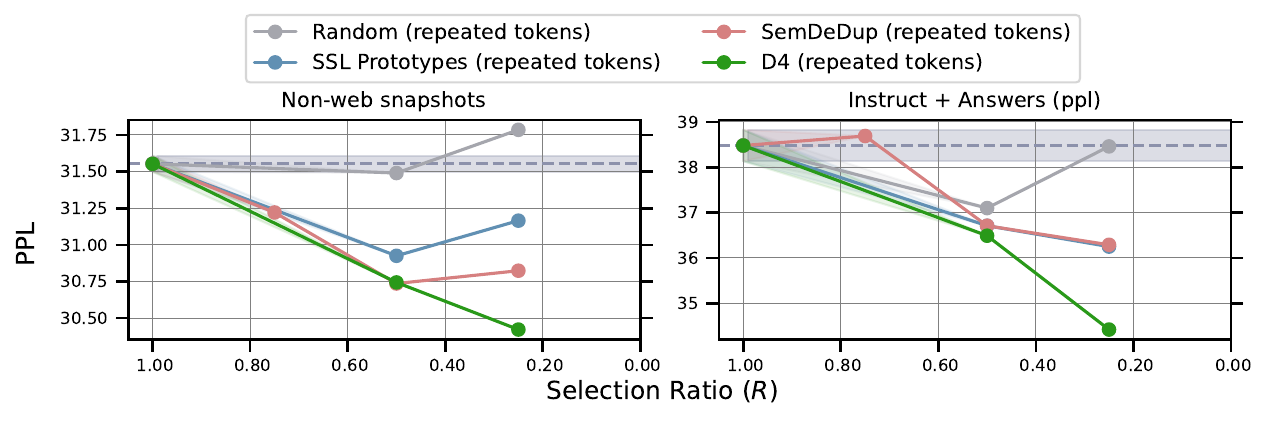}

\caption{Comparison data selection methods when repeating data at the 125M, 3B token budget scale. The x-axis is data selection ratio $R$, and the y-axis is average perplexity on validation sets. We observe that selecting data to repeat via D4 outperforms other data selection methods, especially at low selection ratios $R$ (note that low selection ratios in the fixed-data regime correspond to more epochs).}
\label{fig:appendix_repeated_data_across_data_selection}
\end{center}
\end{figure}

\textbf{Across data selection methods}: We first take the same configuration as Section~\ref{sec:results_repeated_data}, where we have a starting source dataset of 40B tokens, use each of our data selection methods with $R = 0.25$ to select a subset of documents, and repeat over these documents until we reach the target token budget of 40B tokens. Note that this is at the 1.3B model scale. In Figure~\ref{fig:appendix_repeated_data_new_tokens_always_better_non_web_snapshots} we see that repeating data selected by both SemDeDup and SSL prototypes also outperforms randomly selecting new data. However, we quickly notice that for \textit{fixed} data selection strategy (e.g. \textit{fixed} column in Figure~\ref{fig:appendix_repeated_data_new_tokens_always_better_non_web_snapshots}), repeating tokens either outperforms or matched selecting new tokens. In other words: cleverly repeating tokens can outperform randomly selecting new tokens, but if we fix the data selection strategy (random, SemDeDup, SSL prototypes, or D4) then it is usually preferable to select new tokens. We also note in Figure~\ref{fig:appendix_repeated_data_across_data_selection} that D4 outperforms other methods, although by a smaller margin than in the fixed-compute regime.

\begin{figure}[h]
\begin{center}
\includegraphics[width = 1.0\textwidth]{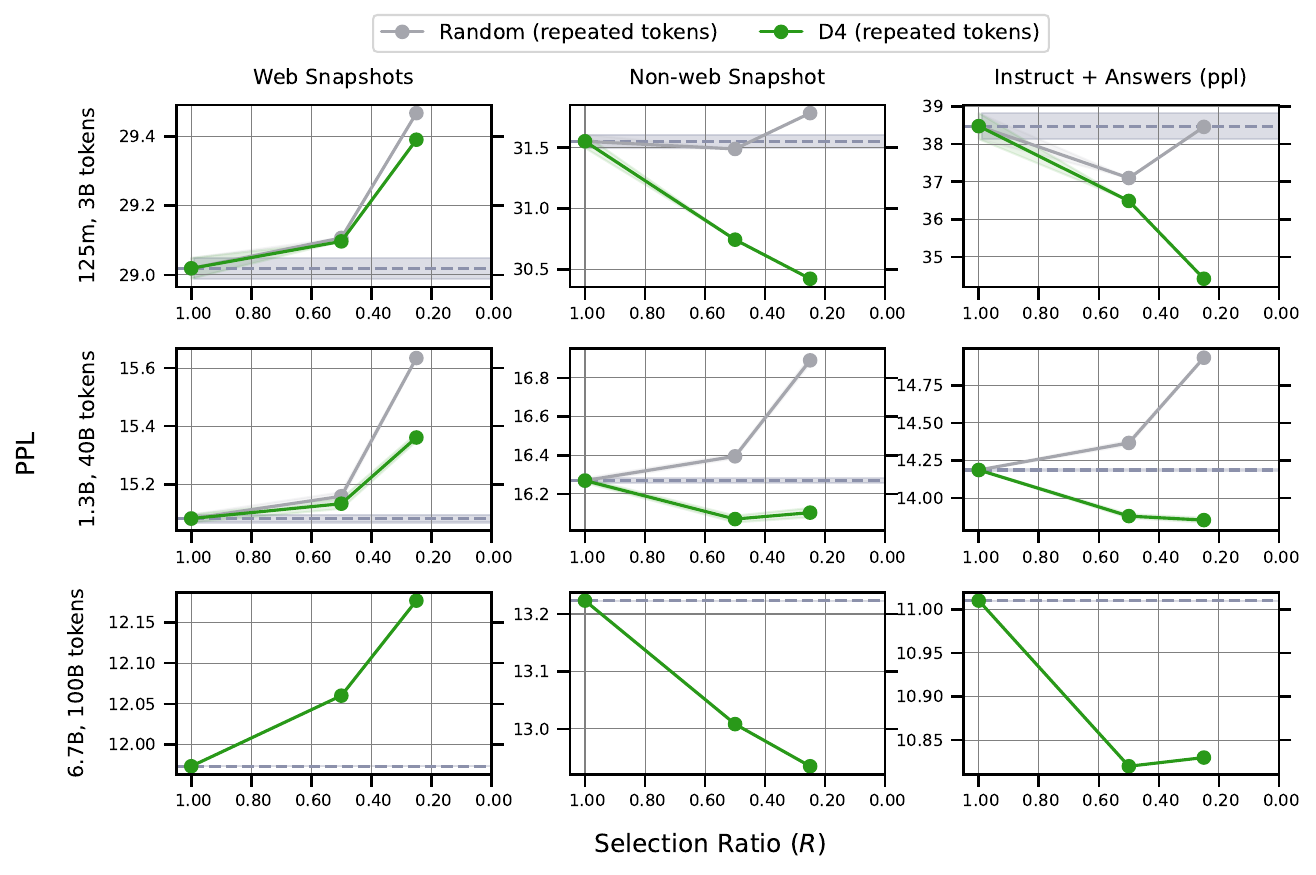}

\caption{Comparison of repeating tokens with D4 (pink line), randomly selecting new tokens (horizontal dashed gray line), and randomly repeating data (gray line). We see across model scales (top: 125M trained on 3B tokens; middle: 1.3B trained on 40B tokens; bottom: 6.7B trained on 100B tokens) and data selection ratios, repeating data selected by D4 outperforms randomly selecting new data.}
\label{fig:appendix_repeated_data_across_model_scale}
\end{center}
\end{figure}

\textbf{Across model scale and data selection ratio}: We fix our data selection strategy as D4 as done in Section~\ref{sec:results_repeated_data}, but attempt repeating tokens across 3 model scales (125M, 1.3B, and 6.7B), and across data selection ratios ($R = 0.5$ and $R = 0.25$). We see in Figure~\ref{fig:appendix_repeated_data_across_model_scale} that repeating data with D4 outperforms randomly selecting new tokens across all model scales and choice of $R$. 

\begin{figure}[h]
\begin{center}
\includegraphics[width = 1.0\textwidth]{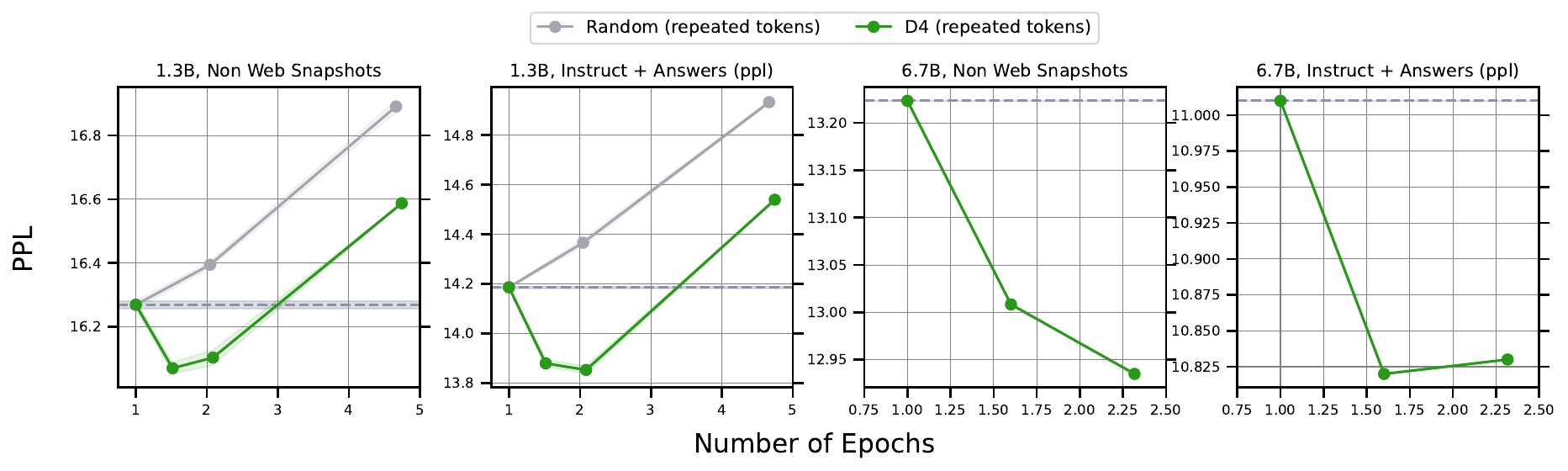}

\caption{Comparison of repeating tokens with D4 (pink line), randomly selecting new tokens (horizontal dashed gray line), and randomly repeating data (gray line). We see with different epoch numbers. The y-axis denotes perplexity, and x-axis denotes number of epochs.}
\label{fig:appendix_repeated_data_across_model_scale_number_epochs}
\end{center}
\end{figure}

We note that for fixed $R$, different data selection methods will choose subsets of the source dataset that contain different amounts of tokens. This means that different data selection methods will epoch a different number of times. For example, for a 1.3B OPT model 40B token budget training run, if randomly repeating data with $R = 0.25$ chooses a subset with 10B tokens and D4 with $R = 0.25$ chooses a subset with 15B tokens, then the random run will epoch 4 times while the D4 run will epoch 2.67 times. To show this more clearly, we plot 1.3B and 6.7B repeated data runs with the x-axis changed to number of epochs in Figure~\ref{fig:appendix_repeated_data_across_model_scale_number_epochs}. We see that up to roughly 2 epochs of data chosen with D4 significantly outperforms randomly selected new data; however, close to 5 epochs leads to worse performance. 

\subsection{Choice of Embedding Space}
\label{sec:appendix_choice_of_embedding_space}
All data selection methods we employ rely heavily on the quality of the underlying embedding space. We qualitatively analyzed the embedding produced by the last-token last-layer OPT 125M model and observed a bias towards end-of-document format. For example, if documents all end with an email or a standard phrase ("Buy our product today!"), then these documents would be clustered together. This likely helps detect templates (since templates tend to end their text in very similar ways), but has clear pitfalls — for example, if we took thousands of wikipedia articles about unrelated topics and appended the same email at the end of each article, they might be clustered together.

Motivated by this, we briefly experiment with different embedding spaces and discuss our results in this section.

\subsubsection{SentenceTransformer models}
BERT embeddings have generally been used to accomplish various NLP tasks, because BERT (unlike GPT/OPT) is able to attend to every token in the input when producing an embedding (BERT is a encoder-decoder model, while OPT/GPT are decoder only). While there are numerous BERT-style models available, we hoped to achieve an embedding space that focused on semantic similarity. Thus, we opted to use the widely popular SentenceTransformer models \footnote{\url{https://www.sbert.net/docs/pretrained_models.html}}, which are BERT-style models finetund specifically >1B text similarity pairs. We choose the top model on the SentenceTransformer leaderboard (all-mpnet-base-v2) and the smallest well-performing model (all-Mini-LM-v6). Note that these models have max context length of 256 and 384 (respectively), and we stuck with the SentenceTransformer default of truncating inputs to fit the max sequence length (i.e. these embeddings only consider the beginning of documents).

\begin{figure}[H]
\begin{center}
\includegraphics[width = 1.0\textwidth]{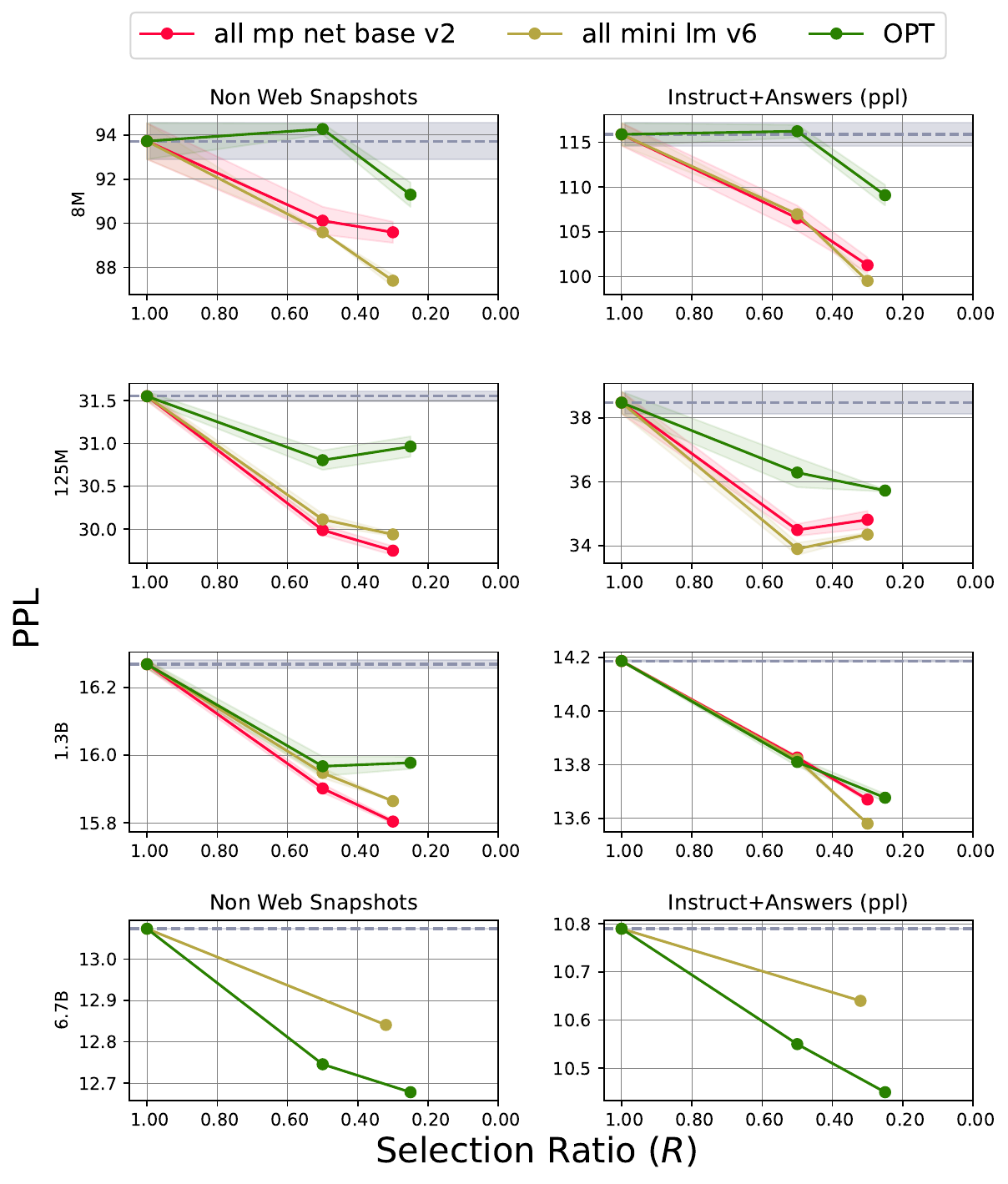}

\caption{Perplexity (y-axis) versus selection ratio $R$ (x-axis) for different embedding spaces, when selecting data via D4. Across different 8m (top), 125m (middle) and 1.3b (bottom) model scales, we see that the SentenceTransformer embedding spaces outperform the OPT embedding space, but at the 6.7b model scale, we see that the OPT embedding space begins outperforming the all Mini LM v6 embedding space. We were unable to run an "all-mp-net-base-v2" 6.7b experiment due to compute restrictions, but we note that the difference between "all-mini-lm-v6" and "all-mp-net-base-v2" across model scales and selection ratios is generally small, so we expect the OPT embedding space to outperform the "all-mp-net-base-v2" at the 6.7b scale.}
\label{fig:appendix_diff_embedding_spaces_ppl}
\end{center}
\end{figure}

\begin{figure}[H]
\begin{center}
\includegraphics[width = 1.0\textwidth]{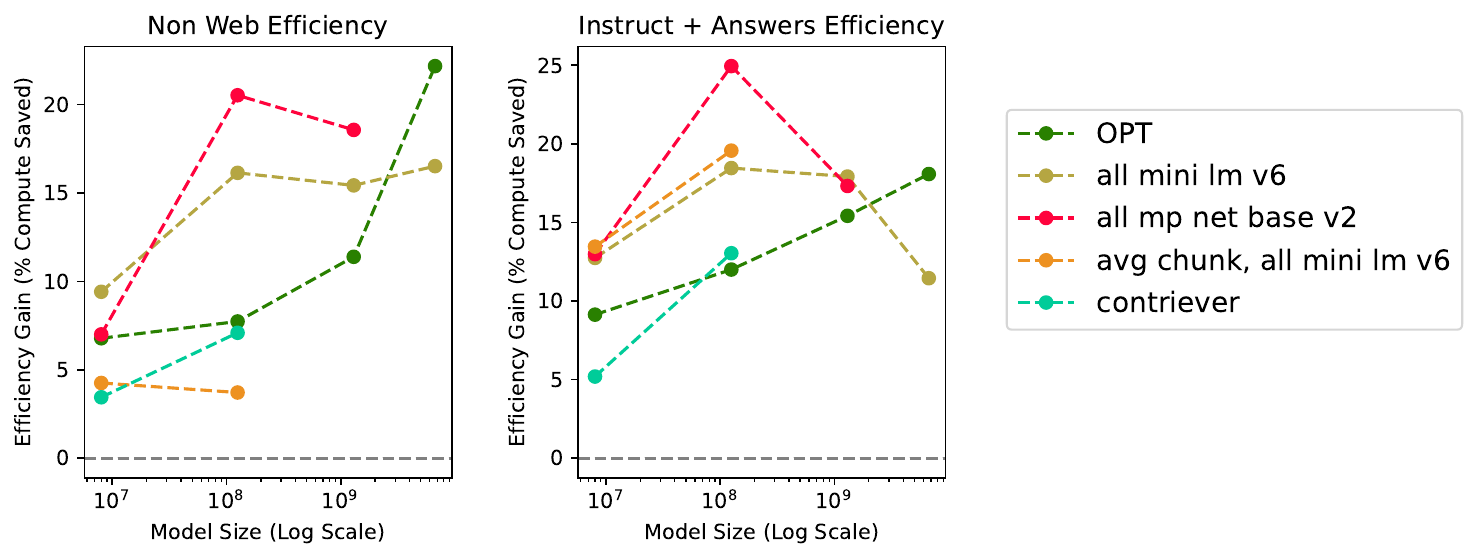}

\caption{Comparison of naive efficiency for different embedding spaces, when using D4 as the data selection strategy. Similar to Figure~\ref{fig:appendix_diff_embedding_spaces_ppl}, we see that all-mini-LM-v6 outperforms the OPT embedding space at small scale, but not at large (6.7b) model scale.}
\label{fig:appendix_diff_embedding_spaces_efficiency}
\end{center}
\end{figure}

We observe, in Figure~\ref{fig:appendix_diff_embedding_spaces_ppl} that at small model scales, sentence transformer embedding spaces outperforms the OPT embedding space. Given these initial results, we took our most overall-all efficient embedding space at the 1.3b model scale ("all-mini-lm-v6") and ran a 6.7b training run with it. Surprisingly, we observed that at larger model scale, the OPT embedding space outperforms the "all-mini-LM-v6" embedding space. Given that the difference between "all-mini-LM-v6" and "all-mp-net-base-v2" is generally small (see Figure~\ref{fig:appendix_diff_embedding_spaces_ppl}), we also expect the OPT embedding space to beat "all-mpnet-base-v2" at the 6.7b, although we were not able to complete this run due to compute restrictions. We see the same trend when we consider overall and naive efficiency of using D4 with different embedding spaces in Figure~\ref{fig:appendix_diff_embedding_spaces_efficiency}.

In an effort to understand why SentenceTransformer embedding spaces perform worse at larger model scales, we qualitatively analyze the clusterings with each SentenceTransformer embedding space. We find that using D4 with "all-mp-net-base-v2" and "all-mini-lm-v6" disproportionately prunes long documents. We hypothesize that this is because sentence transformer models are trained and finetuned on actual sentence pairs, which very rarely saturate the max context length of the model. This might result in all "long" documents (or at least any input that is max-context-length size) seem out-of-distribution to the model. We guess that this results in long documents being clustered together, and therefore disproportionately affected during pruning. This might be especially relevant in domains like Wikipedia articles, where headers and introductions look semantically similar, but the actual content (past the first max-context-length tokens) is very different. 

In an effort to circumvent this problem, we tried two approaches at a small model scale: 

\begin{itemize}
    \item M1\label{def:M1}: Chunking long documents into max-context-length chunks, and averaging all-mini-LM-v6 embeddings across chunks to produce a final document embedding.
    \item M2\label{def:M2}: Using Contriever \citep{izacard2021unsupervised} embeddings, where we chose the Contriever model because it is trained to determine if two sentences are from the same document, and therefore should be agnostic to position within a document.
\end{itemize}

\begin{figure}[H]
\begin{center}
\includegraphics[width = 1.0\textwidth]{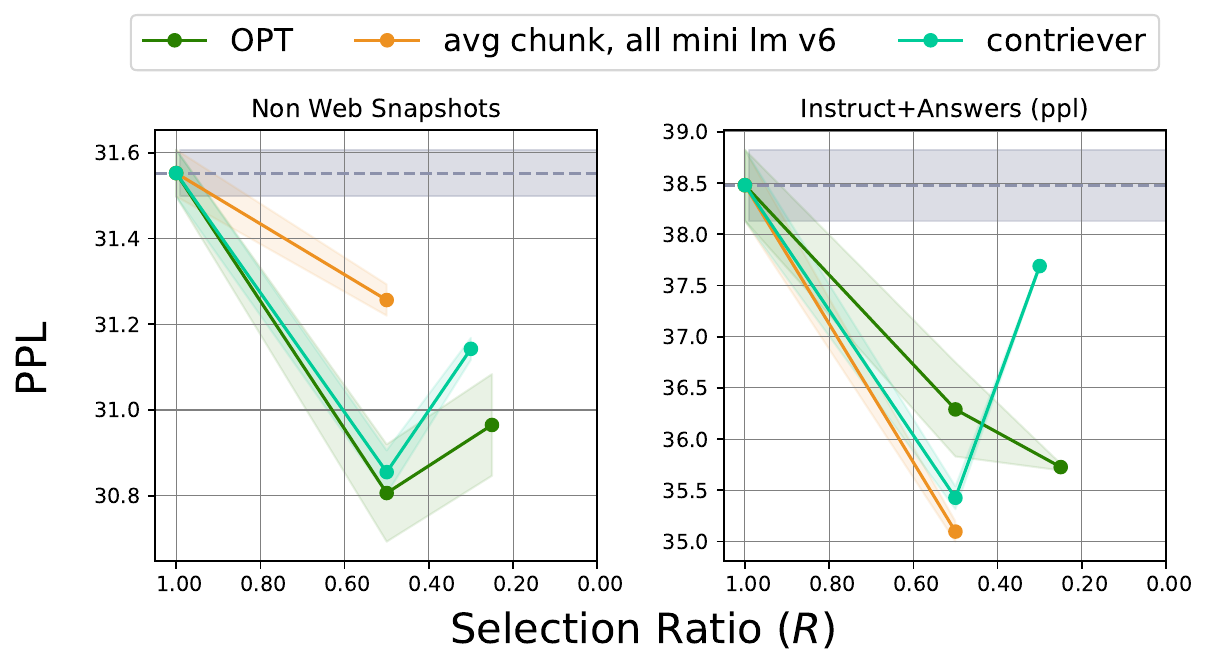}

\caption{Comparison of embedding spaces M1 (averaging embedding of all-mini-LM-v6 across all chunks in a document, where a chunk is defined as 256 tokens) and M2 (embeddings from the Contriever model), with the OPT model embedding space, when using D4 as a the selection strategy. We note that neither embedding space signifigantly outperforms the OPT model embedding space at the 125M scale.}
\label{fig:appendix_diff_embedding_spaces_position_dealing}
\end{center}
\end{figure}

Both in terms of perplexity improvement at the end of training  (see Figure~\ref{fig:appendix_diff_embedding_spaces_position_dealing}) and efficiency (see Figure~\ref{fig:appendix_diff_embedding_spaces_efficiency}) we do not observe a significant difference between the OPT embedding space and embedding spaces M1 and M2 at the small model scale (125 million parameters). We note that M1 and M2 are significantly worse than the all-mp-net-base-v2 and all-mini-LM-v6 at small scales \textbf{and} suffer from the same problem of pruning away long documents (compared to the OPT embedding space), so we expect these models to under-perform the OPT embedding space at the 6.7b scale.

\newpage

\subsection{Replicating Fixed Compute Results on C4}
\label{sec:appendix_small_scale_c4}

\begin{figure}[h]
\begin{center}
\includegraphics[width = 1.0\textwidth]{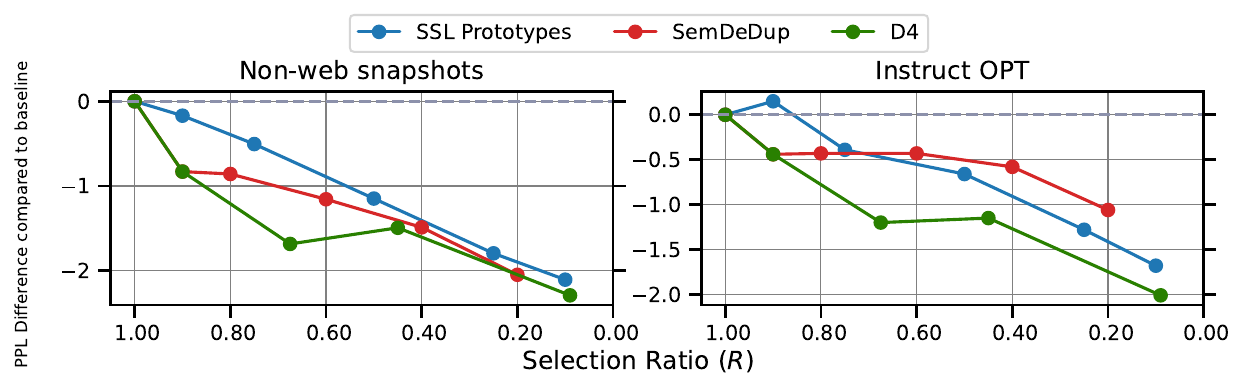}

\caption{Comparison of data selection strategies with the OPT model embedding space, when using D4 as a the selection strategy, when using C4 as the starting training dataset. The x-axis is selectoin ratio $R$, and the y-axis is perplexity difference compared to baseline (the horizontal gray dotted line at 0.0 represents our baseline i.e. when no data selection is done), so \textbf{lower is better}. Notice that D4 and SemDeDup match at 90\%, because we use $R_{dedup} = 0.9$ and varied $R_{proto}$ for this experiment.}
\label{fig:appendix_diff_embedding_spaces_c4}
\end{center}
\end{figure}

In this section, we briefly show our results for comparing data selecting methods at the 125M scale, where the pre-training dataset is the C4 \cite{raffel2020exploring} dataset instead of CC-dedup. We see in Figure~\ref{fig:appendix_diff_embedding_spaces_c4} that D4 generally outperforms other methods. These initial experiments motivates us to try comparing data selection methods on more heavily filtered web-data (i.e. CC-dedup).

\newpage

\subsection{Investigating Duplicate-Driven Clusters}
\label{sec:appendix_examples_duplicate_driven_clusters}

In this subsection, we present a few examples of duplicate-driven clusters, which are clusters that are very dense and near centroids. We find that these clusters tend to be filled with semantic duplicates and/or duplicated text. We generally can find such extreme duplicate-driven clusters by looking at clusters whose standard deviation of cosine distance to cluster centroid is less than 0.03. This is essentially looking at clusters in the lower tail of the empirical CDF in Figure~\ref{fig:cluster_vs_no_recluster_d4} (brown line). We present a few examples of such clusters below:

\begin{table}[h]
\begin{center}
    
\caption{Nearest Neighbors to Cluster Centroid 682}
\label{fig:closest_to_centroid_682}

\begin{tabular}{ l p{10.0cm}  }
\toprule
\textbf{Cosine Distance to Centroid}      
& \textbf{Raw Text}   
\\\midrule
0.03581655 & The USGS (U.S. Geological Survey) publishes a set of the most commonly used topographic maps of the U.S. called US ......... may have differences in elevation and topography, the historic weather at the two separate locations may be different as well. \\\hline0.03584063 & The USGS (U.S. Geological Survey) publishes a set of the most commonly used topographic maps of the U.S. called US ......... may have differences in elevation and topography, the historic weather at the two separate locations may be different as well. \\\hline0.036803484 & The USGS (U.S. Geological Survey) publishes a set of the most commonly used topographic maps of the U.S. called US ......... may have differences in elevation and topography, the historic weather at the two separate locations may be different as well. \\\hline0.037270606 & Search Near Clinton County, OH: Trails National and State Parks City Parks Lakes Lookouts Marinas Historical Sites
The USGS (U.S. Geological ......... may have differences in elevation and topography, the historic weather at the two separate locations may be different as well. \\\hline
\\
\bottomrule
\end{tabular}
\end{center}
\end{table}

\begin{table}[h]
\begin{center}
\caption{Nearest Neighbors to Cluster Centroid 975}
\label{fig:closest_to_centroid_975}

\begin{tabular}{ l p{10.0cm}  }
\toprule
\textbf{Cosine Distance to Centroid}      
& \textbf{Raw Text}   
\\\midrule
0.011662006 & The American Way, Inc.
The American Way, Inc. is a suspended Californian business entity incorporated 19th August 1949. is listed as ......... for bulk data downloadsI want to request the removal of a page on your websiteI want to contact California Explore \\\hline0.012483656 & John St-Amour, Inc.
John St-Amour, Inc. is a suspended Californian business entity incorporated 5th October 1962. is listed as the agent ......... for bulk data downloadsI want to request the removal of a page on your websiteI want to contact California Explore \\\hline0.012564898 & Joseph E. Barbour, Inc.
Joseph E. Barbour, Inc. is a suspended Californian business entity incorporated 27th January 1959. is listed as ......... for bulk data downloadsI want to request the removal of a page on your websiteI want to contact California Explore \\\hline0.012756169 & The Jolly Boys, Inc.
The Jolly Boys, Inc. is a suspended Californian business entity incorporated 4th March 1955. is listed as ......... for bulk data downloadsI want to request the removal of a page on your websiteI want to contact California Explore \\\hline
\\
\bottomrule
\end{tabular}
\end{center}
\end{table}

\begin{table}[h]
\begin{center}
\caption{Nearest Neighbors to Cluster Centroid 10715}
\label{fig:closest_to_centroid_10715}

\begin{tabular}{ l p{10.0cm}  }
\toprule
\textbf{Cosine Distance to Centroid}      
& \textbf{Raw Text}   
\\\midrule
0.035506427 & Search hundreds of travel sites at once for hotel deals at Hotel Olympic
Kornarou Square 44, Heraklion, Greece
34 m Bembo Fountain
262 ......... hundreds of travel sites to help you find and book the hotel deal at Hotel Olympic that suits you best. \\\hline0.036230028 & Search hundreds of travel sites at once for hotel deals at Hotel Estrella del Norte
Juan Hormaechea, s/n, 39195 Isla, Cantabria, ......... travel sites to help you find and book the hotel deal at Hotel Estrella del Norte that suits you best. \\\hline0.036280274 & Search hundreds of travel sites at once for hotel deals at H10 Costa Adeje Palace
Provided by H10 Costa Adeje Palace
Provided ......... travel sites to help you find and book the hotel deal at H10 Costa Adeje Palace that suits you best. \\\hline0.036827266 & Search hundreds of travel sites at once for hotel deals at Hotel Miguel Angel by BlueBay
Calle Miguel Angel 29-31, 28010 ......... sites to help you find and book the hotel deal at Hotel Miguel Angel by BlueBay that suits you best. \\\hline
\\
\bottomrule
\end{tabular}
\end{center}
\end{table}

\begin{table}[h]
\begin{center}

\caption{Random Examples from Cluster 695}
\label{fig:random_examples_cluster_695}

\begin{tabular}{ l p{9.0cm}  }
\toprule
\textbf{Cosine Distance to Cluster Centroid}      
& \textbf{Raw Text}   
\\\midrule
0.044178426 & Eastern Florida State College nutritional sciences
Learn about Eastern Florida State College nutritional sciences, and registering for electives. Which college degrees ......... System (IPEDS). If any stats on Hagerstown Community College career planning are incorrect, please contact us with the right data. \\\hline0.056984067 & Albany State University introduction to business
Find info concerning Albany State University introduction to business, and registering for elective discussion sections ......... If any stats on Warren County Community College plant science major are incorrect, please contact us with the right data. \\\hline0.0534693 & Baldwin Wallace University cost per unit
Learn about Baldwin Wallace University cost per unit, submitting required application forms, and follow-up scheduling. ......... (IPEDS). If any stats on San Jose State nursing degree programs are incorrect, please contact us with the right data. \\\hline0.06892538 & Niagara University managerial accounting
Information about Niagara University managerial accounting, and registering for elective lectures. Which college degrees give you the ......... System (IPEDS). If any stats on Midwestern University pharmacy tech program are incorrect, please contact us with the right data. \\\hline0.07246786 & Fanshawe College app download
Learn about Fanshawe College app download, and registering for elective discussion sections and seminars. Which college degrees ......... Data System (IPEDS). If any stats on Stratford University cell biology are incorrect, please contact us with the right data. \\\hline0.07147932 & Standish Maine Licensed Vocational Nurse LVN Jobs
Find out about Standish, ME licensed vocational nurse LVN jobs options. It's a smart ......... (IPEDS). If any stats on William Jewell College medical insurance coding are incorrect, please contact us with the right data. \\\hline
\\
\bottomrule
\end{tabular}
\end{center}
\end{table}

\begin{table}[h]   
\begin{center}

\caption{Random Examples from Cluster 8342}
\label{fig:random_examples_cluster_8342}

\begin{tabular}{ l p{9.0cm}  }
\toprule
\textbf{Cosine Distance to Cluster Centroid}      
& \textbf{Raw Text}   
\\\midrule
0.027729392 & Seenti - Bundi
Seenti Population - Bundi, Rajasthan
Seenti is a medium size village located in Bundi Tehsil of Bundi district, Rajasthan ......... 6 months. Of 186 workers engaged in Main Work, 63 were cultivators (owner or co-owner) while 0 were Agricultural labourer. \\\hline0.036407113 & Kodunaickenpatty pudur - Salem
Kodunaickenpatty pudur Population - Salem, Tamil Nadu
Kodunaickenpatty pudur is a large village located in Omalur Taluka of ......... 6 months. Of 3523 workers engaged in Main Work, 1500 were cultivators (owner or co-owner) while 1533 were Agricultural labourer. \\\hline0.017463684 & Chhotepur - Gurdaspur
Chhotepur Population - Gurdaspur, Punjab
Chhotepur is a medium size village located in Gurdaspur Tehsil of Gurdaspur district, Punjab ......... 6 months. Of 677 workers engaged in Main Work, 123 were cultivators (owner or co-owner) while 142 were Agricultural labourer. \\\hline0.02616191 & Maksudanpur - Azamgarh
Maksudanpur Population - Azamgarh, Uttar Pradesh
Maksudanpur is a small village located in Sagri Tehsil of Azamgarh district, Uttar ......... 6 months. Of 22 workers engaged in Main Work, 14 were cultivators (owner or co-owner) while 0 were Agricultural labourer. \\\hline0.028420448 & Karambavane - Ratnagiri
Karambavane Population - Ratnagiri, Maharashtra
Karambavane is a medium size village located in Chiplun Taluka of Ratnagiri district, Maharashtra ......... 6 months. Of 444 workers engaged in Main Work, 116 were cultivators (owner or co-owner) while 214 were Agricultural labourer. \\\hline0.037917078 & Barda - Purba Medinipur
Barda Population - Purba Medinipur, West Bengal
Barda is a large village located in Egra - I Block ......... 6 months. Of 1182 workers engaged in Main Work, 278 were cultivators (owner or co-owner) while 252 were Agricultural labourer. \\\hline
\\
\bottomrule
\end{tabular}
\end{center}
\end{table}

\begin{table}[h]
\begin{center}

        \caption{Nearest Neighbors to random validation point in C4}
        \label{fig:appendix_nn_to_valid_c4}
        \begin{tabular}{ l p{10.0cm}  }
            \toprule
    \textbf{Cosine Distance}      
    & \textbf{Raw Text}   
    \\\midrule
    0.0(original validation text) & Offers two child care opportunities to Charles County citizens— the Port Tobacco Onsite Child Care Program and the Before and After School Child Care Program (BASCC).
Supports parents through home visits to first time parents and by helping them search for child care, find resources for a child with social, emotional . . . . . . . . Special needs kids. Free to look, a fee to contact the providers.
Hotline is staffed by highly-trained and friendly Child Care Consumer Education Specialists who offer both parents and providers invaluable information about child care, and referrals to local Child Care Resource and Referral agencies where they can receive individualized assistance. \\\hline0.12867724895477295 & Child Care Options is a program of Options Community Services , a non-profit registered charity dedicated to making a difference in the South Fraser Region. Options is committed to empowering individuals, supporting families and promoting community health. Funding for Child Care Options is provided through British Columbia’s Ministry of Children . . . . . . . . Rock.
Child Care Options links families and child care providers in the communities of Delta, Surrey and White Rock by offering free consultation, support and child care referral services and subsidy support to parents seeking child care. Child care providers are supported through information, outreach, resource library, networking, and learning opportunities. \\\hline0.15080827474594116 & Below are links to child development resources, both from within the department and from external sources.
Child Development Division Publications
Publications that can help you will help you follow your child's development (from birth to age five) so you can identify and address any issues early on.
Resources to help you understand children's . . . . . . . . families to local resources and services. Specialists are available from 9 AM to 6 PM Monday – Friday. Services are confidential. Caregivers can also visit http://www.helpmegrowvt.org/families.html to learn more about child development, discover developmental tips, and watch videos demonstrating children’s developmental milestones (click a button to choose your child’s age). \\\hline0.15738284587860107 & National Domestic Violence Hotlines
Programs that provide immediate assistance for women and men who have experienced domestic abuse which may include steps to ensure the person's safety; short-term emotional support; assistance with shelter; legal information and advocacy; referrals for medical treatment; ongoing counseling and/or group support; and other related services. Hotline . . . . . . . . RP-1500.1400-200)
www.thehotline.org/
Toll Free Phone: 800-799-SAFE
URL: https://www.thehotline.org/
Eligibility: Anyone affected by relationship abuse.
Services Provided: Available 24/7/365 via phone, TTY, and chat. Provides lifesaving tools and immediate support to enable victims to find safety and live lives free of abuse. Highly trained, experienced advocates offer support, crisis intervention, education, safety planning, and referral services. \\\hline
    \\
            \bottomrule
        \end{tabular}
    \end{center}
    \end{table}

\begin{table}[h]
\begin{center}
        \caption{Nearest Neighbors to random validation point in USPTO}
        \label{fig:appendix_nn_to_valid_uspto}
    
        \begin{tabular}{ l p{10.0cm}  }
            \toprule
    \textbf{Cosine Distance}      
    & \textbf{Raw Text}   
    \\\midrule
    0.0(original validation text) & SONET (Synchronous Optical NETwork) is a North American transmission standard for optical communication systems. SDH (Synchronous Digital Hierarchy), a European transmission standard, is a minor variant of SONET.
SONET defines a hierarchy of electrical signals referred to as Synchronous Transport Signals (STS). The STS hierarchy is built upon a basic signal . . . . . . . . the corresponding row and column numbers may include up to 18 comparison operations, which are onerous to implement, for example, in terms of the required logic circuitry. This problem is exacerbated at the upper levels of the STS hierarchy, where processing of multiple pointer values per data frame is performed. \\\hline0.1998944878578186 & US20080109728A1 - Methods and Systems for Effecting Video Transitions Represented By Bitmaps - Google Patents
Methods and Systems for Effecting Video Transitions Represented By Bitmaps Download PDF
David Maymudes
Multi-media project editing methods and systems are described. In one embodiment, a project editing system comprises a multi-media editing application that is configured to . . . . . . . . synchronization models for multimedia data
US20120206653A1 (en) 2012-08-16 Efficient Media Processing
US6658477B1 (en) 2003-12-02 Improving the control of streaming data through multiple processing modules
US6212574B1 (en) 2001-04-03 User mode proxy of kernel mode operations in a computer operating system
US7752548B2 (en) 2010-07-06 Features such as titles, transitions, and/or effects which vary according to positions \\\hline0.21122217178344727 & Both the Ethernet II and IEEE 802.3 standards define the minimum frame size as 64 bytes and the maximum as 1518 bytes. This includes all bytes from the Destination MAC Address field through the Frame Check Sequence (FCS) field. The Preamble and Start Frame Delimiter fields are not included when . . . . . . . . frame. Dropped frames are likely to be the result of collisions or other unwanted signals and are therefore considered invalid.
At the data link layer the frame structure is nearly identical. At the physical layer different versions of Ethernet vary in their method for detecting and placing data on the media. \\\hline0.2133803367614746 & A byte is a group of bits, usually eight. As memory capacities increase, the capacity of chip cards is often quoted in bytes rather than in bits as in the past. \\\hline
    \\
            \bottomrule
        \end{tabular}
    \end{center}
    \end{table}